\documentclass[a4paper,fleqn]{cas-sc}

\usepackage[numbers]{natbib}
\usepackage{booktabs}

\begin{document}

\let\WriteBookmarks\relax
\def\floatpagepagefraction{1}
\def\textpagefraction{.001}
\shorttitle{Reliability-Centered Evaluation of Longitudinal CT Lesion-Size Forecasting}

\shortauthors{Kong et al.}

\title{Reliability-Centered Evaluation of Sparse Longitudinal CT Lesion-Size Forecasting
with Conformal Interval Calibration and Gompertz-Inspired Regularization}


\author[1]{Lingfei Kong\corref{cor1}}

\cortext[cor1]{Corresponding author}

\ead{lingfei.kong@vanderbilt.edu}

\affiliation[1]{
    organization={Department of Mathematics, Vanderbilt University},
    city={Nashville},
    state={Tennessee},
    country={United States}
}


%


%


\begin{abstract}
Sparse longitudinal CT follow-up limits lesion-size forecasting when only a few
prior observations are available. We constructed a five-visit DLT-derived same-lesion trajectory benchmark from DeepLesion and Deep Lesion Tracker (DLT) by linking longitudinal
lesion matches and applying patient- and scan-level quality control, yielding
205 trajectories from 129 patients. We compared an exploratory conventional
sparse-to-final analysis with a primary fixed visit-index horizon design that
predicted the common log change from $T_3$ to $T_4$ while progressively adding
earlier observations. We evaluated predictive accuracy, uncertainty
reliability, post-hoc conformal interval calibration, subgroup performance,
and Gompertz-inspired trajectory regularization. The evaluated methods showed
partially overlapping point-prediction accuracy but distinct uncertainty
behavior. Under the fixed visit-index horizon design, mean held-out RMSE across ten training seeds was 0.4726, 0.4305, 0.4499, and 0.4513 for
$m=1,2,3,4$, respectively, indicating the lowest mean RMSE at $m=2$; additional
history at $m=3$ and $m=4$ did not improve RMSE. At $m=4$, raw Cohort-Level Feature GP
coverage was near the 95\% nominal level, whereas MC Dropout, Deep Ensemble,
and residual-scale intervals were conservative; mixture-aware PIT errors further
distinguished probabilistic calibration. Patient-level conformal calibration
generally produced near-nominal or conservative coverage at the cost of wider
intervals. Patient-grouped development cross-validation selected
$\lambda^*=0$ for the Gompertz-inspired term. A global population reference
frequently opposed lesion-level change directions, and the one-time held-out
comparison did not show a uniform accuracy or reliability benefit. Prediction
difficulty also varied across anatomical subgroups. Overall, additional
historical observations provided limited predictive benefit once the visit-index
prediction horizon was controlled, while predictive accuracy, uncertainty
reliability, and trajectory consistency did not necessarily improve together.
These measures capture distinct aspects of model performance and should be
evaluated jointly in sparse longitudinal imaging.

\end{abstract}


\maketitle

\section{Introduction}
\label{sec:introduction}

Longitudinal CT imaging is widely used to monitor lesion progression,
treatment response, and individualized patient management
\cite{eisenhauer2009recist,zhang2020spatiotemporal}. Changes in lesion size
over time provide useful information about disease trajectory, yet longitudinal
prediction remains difficult because follow-up imaging is often sparse and
irregularly sampled
\cite{zhang2020spatiotemporal,che2018recurrent,rubanova2019latent}.

Prediction accuracy alone is not sufficient for this setting. In clinical use,
a model should also indicate when its predictions are uncertain
\cite{begoli2019need,gawlikowski2023survey}. Models with similar RMSE can
produce very different prediction intervals, and poorly calibrated uncertainty
may limit the practical value of otherwise accurate predictions. This motivates
evaluating sparse longitudinal prediction jointly in terms of point accuracy
and uncertainty reliability
\cite{ovadia2019trust,gawlikowski2023survey}.

\subsection{Related Work and Research Gaps}
\label{subsec:related_work}

Prior work relevant to this study can be grouped into three areas.

\textbf{Benchmark gap.}
Public CT resources such as DeepLesion have enabled large-scale lesion
analysis from routine examinations \cite{yan2018deeplesion}, but most work has
focused on lesion detection, characterization, or other cross-sectional tasks.
Longitudinal prediction studies remain less common and have often used smaller
cohorts; for example, Zhang et al. evaluated spatio-temporal tumor-growth
prediction using data from 33 patients \cite{zhang2020spatiotemporal}. Deep
Lesion Tracker (DLT) introduced longitudinal lesion matching across DeepLesion
scans \cite{cai2021deep}, but these resources have not been widely used to
establish a standardized benchmark for sparse multi-visit same-lesion
prediction.

\textbf{Reliability gap.}
Predictive uncertainty can be estimated using methods such as MC Dropout
\cite{gal2016dropout}, deep ensembles
\cite{lakshminarayanan2017simple}, Bayesian approximations
\cite{mackay1992practical,maddox2019simple}, and conformal prediction
\cite{vovk2005algorithmic,lei2018distribution,romano2019conformalized}.
Previous work has also shown that uncertainty estimates can be poorly
calibrated, particularly under challenging data conditions or dataset shift
\cite{guo2017calibration,kuleshov2018accurate,ovadia2019trust}. However,
these methods have been studied much less systematically in sparse
longitudinal CT lesion-size forecasting.

\textbf{Trajectory-modeling gap.}
Mathematical growth models, including Gompertz-type formulations, have long
been used to describe tumor trajectories
\cite{gompertz1825nature,laird1964dynamics,norton1988gompertzian,
simeoni2004predictive}. Their assumptions, however, may not hold uniformly
across heterogeneous lesions and anatomical sites. Data-driven longitudinal
models \cite{zhang2020spatiotemporal,tao2022prediction} and physics-informed
learning \cite{raissi2019physics,karniadakis2021physics} provide alternative
ways to incorporate temporal structure. More recently, reaction--diffusion
modeling has been combined with deep learning to predict future pulmonary
nodule morphology from longitudinal CT \cite{cai2026pulmonary}. However, the
effect of using a Gompertz-inspired trajectory constraint as a training-time
regularizer on predictive uncertainty and calibration remains insufficiently
characterized.

Taken together, few studies have systematically evaluated these dimensions jointly under a controlled sparse longitudinal CT forecasting setting.

\subsection{Study Overview and Contributions}
\label{subsec:contributions}

To address these gaps, we constructed a five-visit DLT-derived longitudinal lesion trajectory benchmark from
DeepLesion \cite{yan2018deeplesion} and Deep Lesion Tracker
\cite{cai2021deep}. After patient- and scan-level quality control, the final
cohort contained 205 trajectories from 129 patients.

We compared an exploratory conventional sparse-to-final analysis with a
primary fixed visit-index horizon recent-history analysis that held $T_3$ as the most
recent observation and $T_4$ as the forecast endpoint while progressively
earlier observations were added. Deterministic prediction, Cohort-Level Feature GP, MC Dropout
\cite{gal2016dropout}, Deep Ensemble \cite{lakshminarayanan2017simple}, and a
Gaussian residual-scale baseline
were compared using point-prediction and uncertainty metrics. We further
examined whether Gompertz-inspired trajectory regularization altered predictive
accuracy or calibrated uncertainty, while treating residual reduction as a
property of the regularized objective rather than evidence of a universal
biological growth law.

\begin{figure}[pos=h]
    \centering
    \includegraphics[width=0.5\linewidth]{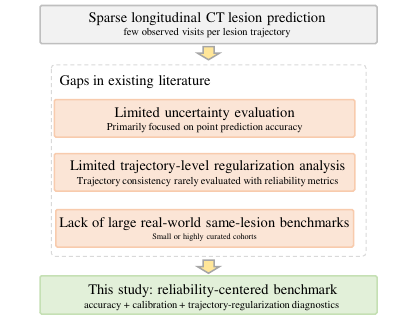}
    \caption{Motivation and research gaps in sparse longitudinal CT lesion
    prediction. The study addresses limitations in benchmark scale,
    uncertainty reliability, and trajectory-level regularization through a
    reliability-centered evaluation framework.}
    \label{fig:motivation}
\end{figure}

Figure~\ref{fig:motivation} summarizes the motivation and study design.

The main contributions of this study are fourfold:

\begin{itemize}

\item \textbf{Benchmark construction.}
We establish a patient-level, five-visit same-lesion CT forecasting benchmark
from DeepLesion-DLT, with explicit patient- and scan-level quality control and
leakage-aware cohort construction.

\item \textbf{Controlled longitudinal evaluation.}
We use a fixed visit-index horizon recent-history design to isolate the incremental
contribution of progressively older observations while holding the forecast endpoint and most recent observed visit fixed.

\item \textbf{Reliability-centered evaluation.}
We evaluate point prediction and predictive uncertainty jointly using coverage,
interval width, proper interval scores, distributional scores, calibration, and
uncertainty--error alignment, rather than relying on prediction error alone.

\item \textbf{Structural-consistency analysis.}
We examine whether Gompertz-inspired trajectory regularization improves
predictive reliability or instead introduces a trade-off between agreement
with a fixed trajectory reference and forecasting performance.

\end{itemize}


\section{Materials and Methods}
\label{sec:methods}


\subsection{Data Sources}
\label{subsec:data_sources}

This study used DeepLesion \cite{yan2018deeplesion} and DeepLesion Tracking
(DLT) \cite{cai2021deep} to construct a longitudinal CT lesion benchmark.
DeepLesion provides lesion annotations, RECIST-style long- and short-axis
measurements, and patient/scan metadata, while DLT provides source--target
lesion matches across longitudinal CT scans. DeepLesion measurements were used
to quantify lesion size, and the released DLT matches were used to construct
DLT-derived longitudinal lesion trajectories without additional image-based
rematching.

\begin{table}[pos=h]
\centering
\caption{Datasets and metadata used in the DeepLesion-DLT benchmark.}
\label{tab:datasets}
\begin{tabular}{p{0.18\linewidth} p{0.39\linewidth} p{0.35\linewidth}}
\toprule
\textbf{Data component} &
\textbf{Information provided} &
\textbf{Use in this study} \\
\midrule

DeepLesion &
RECIST measurements and patient/scan metadata &
Quantify lesion size at each visit \\

DLT &
Source--target lesion matches across CT scans &
Construct DLT-derived longitudinal lesion trajectories \\

LesaNet / metadata &
Body-site and lesion-type labels &
Define anatomical subgroups \\

Patient/scan QC &
PatientID, StudyID, and ScanID &
Verify patient consistency and distinct CT visits \\

\bottomrule
\end{tabular}
\end{table}

\begin{figure}[pos=h]
\centering
\includegraphics[width=0.98\linewidth]{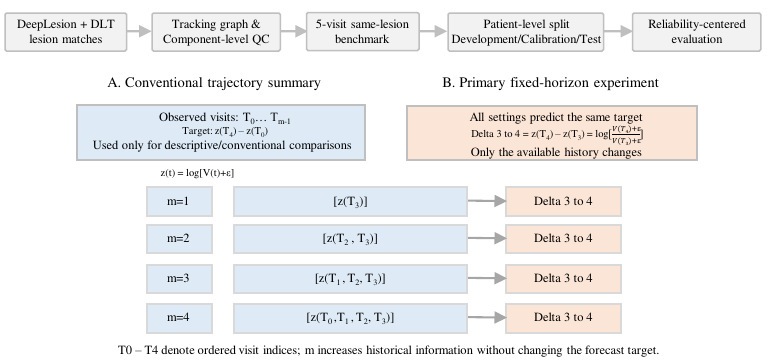}
\caption{Construction of the DeepLesion-DLT longitudinal benchmark and prediction tasks. DLT lesion matches were assembled into longitudinal lesion trajectories, filtered by patient- and scan-level quality control, and split at the patient level. The primary fixed visit-index horizon uses absolute log-volume histories ending at $T_3$ and predicts the common target $\Delta_{3\rightarrow4}=z(T_4)-z(T_3)$ for every history length $m$; increasing $m$ adds earlier observations without changing the target.}
\label{fig:benchmark_construction}
\end{figure}

Figure~\ref{fig:benchmark_construction} summarizes benchmark construction and
the subsequent prediction workflow.

\subsection{Longitudinal Trajectory Construction}
\label{subsec:trajectory_construction}

Each lesion observation was represented as a graph node and each DLT
source--target match as an edge, defining

\begin{equation}
    G=(V,E),
    \label{eq:graph}
\end{equation}

where $v\in V$ denotes a lesion observation from one CT scan and
$(v_i,v_j)\in E$ denotes a DLT-provided longitudinal match.

Connected components were extracted as

\begin{equation}
    \mathcal{C}=\{C_1,C_2,\ldots,C_K\},
    \label{eq:components}
\end{equation}

with each component $C_k\subseteq V$ treated as a candidate same-lesion
trajectory. This construction links sequential DLT matches into multi-visit
trajectories without requiring additional pairwise lesion matching.

When a component contained multiple candidate nodes from the same scan, the
main construction retained a node with a valid RECIST measurement and resolved
remaining ties by an immutable node identifier. This deterministic rule used
neither measurements from later visits, the fifth-visit forecast endpoint, nor
trajectory smoothness. A prespecified strict sensitivity analysis excluded any
component containing a scan with zero or more than one valid RECIST candidate,
thereby assessing robustness to candidate-node ambiguity.


\subsection{Patient- and Scan-Level Quality Control}
\label{subsec:quality_control}

PatientID, StudyID, and ScanID were parsed from the DeepLesion/DLT file names.
A candidate component was retained only when all of its lesion nodes belonged
to the same patient:

\begin{equation}
    \left|
    \left\{
    \operatorname{PatientID}(v):v\in C_k
    \right\}
    \right|=1.
    \label{eq:patient_qc}
\end{equation}

To ensure that repeated nodes from the same scan were not counted as separate
follow-up visits, retained components were also required to contain at least
five unique scans:

\begin{equation}
    \left|
    \left\{
    \operatorname{ScanID}(v):v\in C_k
    \right\}
    \right|\geq5.
    \label{eq:scan_qc}
\end{equation}

When multiple candidate nodes from the same component occurred within one
scan, priority was given to nodes with complete RECIST measurements and any
remaining tie was resolved by an immutable node identifier. The graph audit
explicitly counted branching nodes, repeated pair annotations, scans with
multiple candidates, and ambiguity exclusions. No future lesion size,
trajectory smoothness, or target-visit measurement was used for candidate
selection. The strict sensitivity cohort excluded components with ambiguous
valid candidates; in the present data it was identical to the main cohort.
A prespecified 50-trajectory manifest was additionally generated for future
image-level manual review; because such review was not available in the
released annotations, no radiologist-verification claim is made here.

We audited the directionality of all released DLT pair records before assigning visit indices. The 7,776 directed records represented 3,888 reciprocal lesion pairs: 3,888 records followed ascending StudyID/ScanID order, 3,888 followed the reverse order, every record had its reciprocal orientation, and all connected endpoints belonged to the same patient. Thus, the DLT source and target fields were treated as reciprocal graph-edge labels rather than chronological designations. After undirected connected-component construction, eligible trajectories were ordered using StudyID and ScanID. Because exact CT acquisition dates were not consistently available, this sequence defines an operational visit index, not verified acquisition chronology or elapsed clinical time. The first five ordered observations were designated \(T_0,T_1,T_2,T_3,T_4\) for the main analysis.


\subsection{Cohort Definition and Data Split}
\label{subsec:data_split}

After trajectory construction and quality control, 205 same-lesion
trajectories from 129 patients were retained. Splitting was performed at the
patient level, so all trajectories from one patient belonged to exactly one
subset. The development set contained 140 trajectories from 84 patients, the
dedicated calibration set contained 27 trajectories from 19 patients, and the
held-out test set contained 38 trajectories from 26 patients
(Table~\ref{tab:data_split}).

Model and hyperparameter selection were confined to patient-grouped five-fold
cross-validation within the development patients. The calibration patients
were used only to construct patient-level conformal intervals, and the test
patients were accessed only for final evaluation. Thus, neither calibration nor
test outcomes contributed to model selection.

\begin{table}[pos=h]
\centering
\caption{Patient-level development/calibration/test split.}
\label{tab:data_split}
\begin{tabular}{lccp{0.38\linewidth}}
\toprule
\textbf{Set} & \textbf{Patients} & \textbf{Trajectories} & \textbf{Purpose} \\
\midrule
Development & 84 & 140 & Model fitting and patient-grouped cross-validation \\
Calibration & 19 & 27 & Patient-level conformal calibration only \\
Test & 26 & 38 & Final evaluation only \\
\bottomrule
\end{tabular}
\end{table}


\subsection{Prediction Task and Target Definition}
\label{sec:prediction-task}

Because DeepLesion-DLT does not provide three-dimensional segmentation masks,
lesion size was represented using a RECIST-based ellipsoid volume proxy
\cite{eisenhauer2009recist}. Let $L(t)$ and $S(t)$ denote the RECIST long and
short axes at visit $t$, respectively. The volume proxy was

\begin{equation}
    V(t)
    =
    \frac{\pi}{6}L(t)S(t)^2.
    \label{eq:volume_proxy}
\end{equation}

This deterministic measure is a lesion-size surrogate derived from two
orthogonal RECIST diameters, not a segmented three-dimensional tumor volume.

For numerical stability, lesion size was represented on the absolute log scale,

\begin{equation}
    z(t)=\log\{V(t)+\epsilon\},
    \label{eq:absolute_log_volume}
\end{equation}

where $\epsilon$ is a numerical stabilizer. All retained lesions had positive
RECIST measurements, so $\epsilon$ was set to zero in the main analysis. The
primary fixed-horizon outcome was the log change from the most recent observed
visit $T_3$ to the common target visit $T_4$:

\begin{equation}
    \Delta_{3\rightarrow4}=z(T_4)-z(T_3)
    =\log\left\{\frac{V(T_4)+\epsilon}{V(T_3)+\epsilon}\right\}.
    \label{eq:t3_t4_log_change}
\end{equation}

Thus, the $m=1$ setting uses the current lesion scale at $T_3$ and does not
implicitly use $T_0$. For neural trajectory models, prediction was carried out
on the absolute $z(T_4)$ scale and translated to $\Delta_{3\rightarrow4}$ by
subtracting $z(T_3)$; all reported fixed-horizon errors were computed on this
same $T_3$-to-$T_4$ log-change outcome.


\subsection{History Depth Analysis}
\label{subsec:analysis-designs}

We considered two complementary analysis designs. The exploratory conventional
sparse-to-final analysis used the first $m$ observed visits to predict the
baseline-relative change $z(T_4)-z(T_0)$: $T_0\rightarrow T_4$ for $m=1$,
$(T_0,T_1)\rightarrow T_4$ for $m=2$, $(T_0,T_1,T_2)\rightarrow T_4$ for
$m=3$, and $(T_0,T_1,T_2,T_3)\rightarrow T_4$ for $m=4$. Because increasing
$m$ changed both history depth and the proximity of the latest observation to
$T_4$, cross-history differences in this analysis were interpreted
descriptively.

The primary fixed visit-index horizon analysis used the common outcome
$\Delta_{3\rightarrow4}$ from Eq.~\eqref{eq:t3_t4_log_change}. Every input
window ended at $T_3$, and progressively earlier absolute log-volume
observations were added: $z(T_3)$ for $m=1$; $(z(T_2),z(T_3))$ for $m=2$;
$(z(T_1),z(T_2),z(T_3))$ for $m=3$; and
$(z(T_0),z(T_1),z(T_2),z(T_3))$ for $m=4$. Equivalently, the models received
the current $T_3$ scale together with the available within-history log changes.
This design held both the latest observation and the one-visit prediction
horizon fixed, so differences across $m$ isolate the incremental value of
progressively older history under the evaluated model.


\subsection{Compared Predictive and UQ Methods}
\label{subsec:compared_methods}

Five prediction and uncertainty-quantification methods were compared.

\paragraph{Deterministic MLP.}
A multilayer perceptron was used as the point-prediction baseline and produced
no native prediction interval.

\paragraph{Cohort-Level Feature GP.}
An exact Cohort-Level Feature GP used standardized recent-history features comprising
relative visit indices, observed absolute log lesion sizes, and the visit-index forecast gap.
Its kernel was the product of a constant kernel and an anisotropic radial-basis
function kernel, with an additive white-noise kernel. Kernel hyperparameters
were fitted using development data only. The posterior mean was used for point
prediction and the posterior standard deviation defined the raw Gaussian 95\%
predictive interval.

\paragraph{MC Dropout.}
MC Dropout \cite{gal2016dropout} retained dropout during inference. Each
stochastic pass $s$ produced both a predictive mean $\mu_s(x)$ and a
heteroscedastic variance $\sigma_s^2(x)$. Its total predictive variance was
\begin{equation}
\widehat{\operatorname{Var}}(Y\mid x)
=
\frac{1}{S}\sum_{s=1}^{S}\sigma_s^2(x)
+
\operatorname{Var}_{s}\{\mu_s(x)\},
\label{eq:mc_total_variance}
\end{equation}
separating aleatoric variation from between-pass epistemic disagreement.

\paragraph{Deep Ensemble.}
Deep Ensemble \cite{lakshminarayanan2017simple} combined $K=5$
independently initialized heteroscedastic neural networks. Its total predictive
variance used the analogous within- and between-model decomposition detailed
in Eq.~\eqref{eq:ensemble_total_variance}. Raw MC Dropout and Deep Ensemble
NLL values were computed from their Gaussian-mixture predictive densities,
rather than from interval width.

\paragraph{Gaussian residual-scale baseline.}
The exact fitted deterministic MLP used in the Deterministic comparator
produced the predictive mean
\begin{equation}
\hat{y}_i=f_{\theta}(x_i).
\end{equation}
No second network was trained for this baseline. Training-partition residuals,
\begin{equation}
r_j=y_j-\hat{y}_j,
\end{equation}
were used to estimate
\begin{equation}
\bar{r}
=
\frac{1}{N_{\mathrm{train}}}
\sum_{j\in\mathcal{D}_{\mathrm{train}}} r_j,
\end{equation}
and
\begin{equation}
\hat{\sigma}_{\mathrm{res}}
=
\sqrt{
\frac{1}{N_{\mathrm{train}}-1}
\sum_{j\in\mathcal{D}_{\mathrm{train}}}
(r_j-\bar{r})^2
}.
\end{equation}

The raw $(1-\alpha)$ prediction interval was
\begin{equation}
\mathcal{I}^{\mathrm{raw}}_i
=
\left[
\hat{y}_i-z_{1-\alpha/2}\hat{\sigma}_{\mathrm{res}},
\;
\hat{y}_i+z_{1-\alpha/2}\hat{\sigma}_{\mathrm{res}}
\right].
\end{equation}
Because the same residual scale was used for all samples, the raw intervals
had constant width before post-hoc calibration. By construction, this baseline
and the Deterministic comparator therefore had identical point predictions and
RMSE within every repeat; only the estimated residual scale and interval output
were added.

\paragraph{Gompertz-inspired trajectory regularization.}
Gompertz-inspired regularization was evaluated as an auxiliary trajectory
constraint motivated by the classical Gompertz formulation
\cite{gompertz1825nature}. A single population-level reference curve was
estimated using training-set longitudinal observations only, and its parameters
were frozen before model fitting, validation, and testing. The exact
training-only estimation procedure, residual, and objective are defined in
Section~\ref{subsubsec:gompertz_regularization}. The residual was treated as a
regularization diagnostic rather than evidence of a universal biological
growth law.


\subsection{Uncertainty Calibration and Evaluation Metrics}
\label{subsec:uq_calibration}

Raw predictive distributions and post-hoc conformal intervals were evaluated
as distinct statistical objects. Raw probabilistic predictions were assessed using predictive-density NLL and a PIT-based probabilistic calibration error. For case $i$, let
\begin{equation}
u_i=\widehat F_i(y_i),
\end{equation}
where $\widehat F_i$ is the model's predictive CDF. For MC Dropout and Deep
Ensemble, $\widehat F_i$ was the equal-weight Gaussian-mixture CDF over
stochastic passes or ensemble members, matching the mixture density used for
NLL. The Cohort-Level Feature GP and Gaussian residual-scale baseline used
their Gaussian predictive CDFs. At grid points
$G=\{0.05,0.10,\ldots,0.95\}$, the empirical PIT CDF was
\begin{equation}
\widehat H(g)=\frac{1}{n}\sum_{i=1}^{n}\mathbb{I}(u_i\le g),
\end{equation}
and the reported PIT calibration error was
\begin{equation}
\mathrm{DCE}_{\mathrm{PIT}}=
\frac{1}{|G|}\sum_{g\in G}|\widehat H(g)-g|.
\end{equation}
Lower values indicate closer agreement with the Uniform$(0,1)$ distribution expected under PIT calibration.
This definition avoids converting mixture-predictive interval width into a
single Gaussian standard deviation. Post-hoc conformal outputs were treated as
set-valued intervals and were assessed with coverage, width, interval score,
and weighted interval score; neither PIT error nor NLL was assigned after
conformalization.

For trajectory $j$ from calibration patient $p$, the absolute-residual score
was $s_{pj}=|y_{pj}-\hat y_{pj}|$. For a method with a native predictive scale
$\hat\sigma_{pj}$, the normalized score was
\begin{equation}
s_{pj}=\frac{|y_{pj}-\hat y_{pj}|}{\hat\sigma_{pj}+\epsilon}.
\label{eq:normalized_calibration_score}
\end{equation}
To respect patient clustering, one score was formed per calibration patient:
\begin{equation}
S_p=\max_j s_{pj}.
\label{eq:patient_conformal_score}
\end{equation}
For $n_{\mathrm{cal}}$ calibration patients and miscoverage level $\alpha$,
the finite-sample order statistic used rank
\begin{equation}
k^{\star}
=
\min\!\left\{
\left\lceil(n_{\mathrm{cal}}+1)(1-\alpha)\right\rceil,
n_{\mathrm{cal}}
\right\},
\qquad
q_{1-\alpha}=S_{(k^{\star})}.
\label{eq:conformal_threshold}
\end{equation}
The calibrated half-width was $q_{1-\alpha}$ for absolute-residual
calibration and $q_{1-\alpha}\hat\sigma(x)$ for normalized calibration.
Calibration was evaluated at nominal 80\%, 90\%, and 95\% coverage. With
$n_{\mathrm{cal}}=19$, their ranks were 16, 18, and 19, respectively. Thus,
at $\alpha=0.05$, $k^{\star}=19$ and the adjustment equals the maximum of
the 19 patient scores. This is the finest available upper-tail order statistic
in this calibration cohort, making the 95\% analysis resolution-limited and
expected to be conservative.

Point prediction was evaluated by RMSE and MAE. For interval
$[\hat\ell_i,\hat u_i]$, coverage and mean width were
\begin{equation}
\mathrm{PICP}
=
\frac{1}{N}\sum_{i=1}^{N}
\mathbb{I}\{y_i\in[\hat\ell_i,\hat u_i]\},
\qquad
\mathrm{MPIW}
=
\frac{1}{N}\sum_{i=1}^{N}(\hat u_i-\hat\ell_i).
\label{eq:picp_mpiw}
\end{equation}
Here PICP is trajectory-level marginal coverage. To align a secondary endpoint
with the patient-level maximum-score construction, we also reported
patient-level simultaneous coverage,
\begin{equation}
\mathrm{PSC}=\frac{1}{N_{\mathrm{test,pat}}}\sum_{p=1}^{N_{\mathrm{test,pat}}}
\mathbb{I}\!\left\{y_{pj}\in[\hat\ell_{pj},\hat u_{pj}]
\;\text{for all }j\in\mathcal{J}_p\right\},
\label{eq:patient_simultaneous_coverage}
\end{equation}
where $\mathcal{J}_p$ indexes all held-out trajectories of test patient $p$.
The central $(1-\alpha)$ interval score was
\begin{equation}
\mathrm{IS}_{\alpha}
=
(\hat u-\hat\ell)
+
\frac{2}{\alpha}(\hat\ell-y)\mathbb{I}(y<\hat\ell)
+
\frac{2}{\alpha}(y-\hat u)\mathbb{I}(y>\hat u),
\label{eq:interval_score}
\end{equation}
and a one-level weighted interval score combined
$|y-\hat y|$ with $\mathrm{IS}_{\alpha}$ using weights $1/2$ and
$\alpha/2$, respectively \cite{gneiting2007strictly}. Lower interval score
and WIS indicate a better coverage--sharpness trade-off.

\begin{table}[htbp]
\centering
\caption{Evaluation metrics and their valid reporting domains.}
\label{tab:evaluation_metrics}
\begin{tabular}{p{0.27\linewidth} p{0.65\linewidth}}
\toprule
\textbf{Metric} & \textbf{Reporting domain} \\
\midrule
RMSE / MAE & Point-prediction error \\
Predictive-density NLL / PIT calibration error & Raw probabilistic predictions only \\
PICP / MPIW & Raw and conformal prediction intervals \\
Interval score / WIS & Proper scoring of raw and conformal intervals \\
Gompertz-style residual & Agreement with the imposed trajectory reference \\
\bottomrule
\end{tabular}
\end{table}

The dedicated calibration patients were not used for model selection, and the
test set remained untouched until final evaluation.


\subsection{Experimental Design}
\label{subsec:experimental_design}

The framework comprised one data-quality audit (Experiment 0) followed by five
predictive and reliability analyses (Experiments 1--5), summarized in
Table~\ref{tab:experimental_design}.

\begin{table}[pos=h]
\centering
\caption{Overview of the experimental design.}
\label{tab:experimental_design}
\begin{tabular}{p{0.09\linewidth} p{0.29\linewidth} p{0.53\linewidth}}
\toprule
\textbf{Exp.} & \textbf{Analysis} & \textbf{Primary objective} \\
\midrule
0 & Data quality audit &
Verify trajectory validity, patient consistency, scan uniqueness, and benchmark composition \\
1 & History Depth and Fixed Visit-index Prediction-Horizon Analysis &
Evaluate performance as observed follow-up increases and examine subgroup behavior \\
2 & UQ method comparison &
Compare deterministic prediction, Cohort-Level Feature GP, MC Dropout, Deep Ensemble, and the Gaussian residual-scale baseline \\
3 & Development-CV Regularization Selection and Population-Reference Diagnostics &
Select $\lambda$ without test access and diagnose population-reference alignment \\
4 & One-time regularized UQ evaluation &
Evaluate the fixed nonzero contrast once on the held-out test set \\
5 & Calibration and subgroup reliability &
Evaluate calibration, subgroup reliability, and uncertainty--error association \\
\bottomrule
\end{tabular}
\end{table}


\subsection{Implementation Details and Statistical Inference}
\label{subsec:implementation}

All neural models were implemented in PyTorch using a sparse-trajectory MLP
encoder. Observed visit indices and absolute log-volume values were padded to
a maximum length of 8 and masked before input to the model:
\begin{equation}
x_i
=
\left[
t_{i,\mathrm{obs}}\odot m_i,\;
z_{i,\mathrm{obs}}\odot m_i,\;
m_i,\;
t_{i,\mathrm{target}}
\right],
\label{eq:model_input}
\end{equation}
where $m_i$ is supplied as an explicit input feature in addition to masking
the padded time and value channels. Consequently, a genuine observation with
$t=0$ or $z=0$ is distinguishable from an unobserved padding slot. The
five-visit benchmark used at most four observed input visits.

\subsubsection{Neural Architecture and Training Settings}
\label{subsubsec:training_settings}

The encoder used two hidden layers with 64 units per layer and SiLU activation.
Dropout was applied only to stochastic UQ models.

\begin{table}[htbp]
\centering
\caption{Model architecture and implementation settings.}
\label{tab:implementation_settings}

\begin{tabular}{ll}
\toprule
Setting & Value \\
\midrule
Implementation & PyTorch \\
Model backbone & Sparse-trajectory MLP encoder \\
Maximum padded length & 8 \\
Hidden layers & 2 \\
Hidden dimension & 64 \\
Activation & SiLU \\
Optimizer & AdamW \\
Learning rate & 0.001 \\
Weight decay & $1\times10^{-5}$ \\
Batch size & 128 \\
Dropout rate & 0.10 \\
MC Dropout samples & 50 \\
Deep Ensemble type & Heteroscedastic, 5 members \\
Split rule & Patient-level development/calibration/test \\
Main split seed & 20260829 \\
Development/calibration/test trajectories & 140 / 27 / 38 \\
Nominal conformal coverage levels & 80\%, 90\%, 95\% \\
\bottomrule
\end{tabular}

\end{table}

Experiment-specific training settings are summarized in
Table~\ref{tab:experiment_training_settings}.

\begin{table}[htbp]
\centering
\caption{Experiment-specific training settings.}
\label{tab:experiment_training_settings}
\begin{tabular}{llll}
\toprule
Experiment & Main comparison & Epochs & Key settings \\
\midrule
Experiment 2 & UQ method comparison & 300 &
MC samples = 50; ensemble size = 5 \\
Experiment 3 & Development-CV weight selection & 300 &
5 patient-grouped folds $\times$ 2 seeds; $\lambda\in\{0,0.1,1,10,100\}$ \\
Experiment 4 & One-time held-out test evaluation & 300 &
MC Dropout UQ; $\lambda^*=0$ primary; best nonzero $\lambda=0.1$ \\
\bottomrule
\end{tabular}
\end{table}

No early stopping was applied; all models were trained for the fixed
number of epochs specified in
Table~\ref{tab:experiment_training_settings}. No additional feature standardization was applied to the neural models beyond
the absolute log transformation defined in
Section~\ref{sec:prediction-task}.
The Cohort-Level Feature GP used the standardized feature representation described in
Section~\ref{subsec:compared_methods}.

\subsubsection{Stochastic UQ Implementation}
\label{subsubsec:stochastic_uq}

For MC Dropout, dropout remained active during inference and $S=50$
stochastic forward passes were used:
\begin{equation}
\hat{y}_i
=
\frac{1}{S}
\sum_{s=1}^{S}
\hat{y}_i^{(s)}.
\label{eq:mc_mean}
\end{equation}
The predictive variance combined the mean within-pass heteroscedastic variance
and the between-pass variance of the predictive means:
\begin{equation}
\hat{\sigma}_i^2
=
\frac{1}{S}
\sum_{s=1}^{S}
\left(\hat{\sigma}_i^{2}\right)^{(s)}
+
\operatorname{Var}_{s}\!\left[\hat{y}_i^{(s)}\right].
\label{eq:mc_total_variance_impl}
\end{equation}
Raw 95\% prediction intervals were constructed using the moment-matched
Gaussian form
$[\hat{y}_i-1.96\hat{\sigma}_i,\,
\hat{y}_i+1.96\hat{\sigma}_i]$.

For Deep Ensemble, the predictive mean was computed across the $K=5$
heteroscedastic ensemble members:
\begin{equation}
\hat{y}_i
=
\frac{1}{K}
\sum_{k=1}^{K}
\hat{y}_{i,k}.
\label{eq:ensemble_mean}
\end{equation}

The total predictive variance combined the mean within-model heteroscedastic
variance and the between-model variance of the predictive means:
\begin{equation}
\hat{\sigma}_i^2
=
\frac{1}{K}
\sum_{k=1}^{K}
\left(
\hat{\sigma}_{i,k}^{2}
+
\hat{y}_{i,k}^{2}
\right)
-
\hat{y}_i^{2}.
\label{eq:ensemble_total_variance}
\end{equation}

Thus, the first component reflects the average sample-dependent uncertainty
predicted by the ensemble members, whereas variation among
$\hat{y}_{i,k}$ reflects disagreement across independently initialized
models. Each ensemble member was trained using a heteroscedastic Gaussian
negative log-likelihood to jointly estimate the predictive mean and variance.
Raw 95\% prediction intervals were then constructed as

\begin{equation}
\left[
\hat{y}_i-1.96\,\hat{\sigma}_i,\;
\hat{y}_i+1.96\,\hat{\sigma}_i
\right].
\label{eq:ensemble_raw_interval}
\end{equation}

\subsubsection{Gompertz-Inspired Trajectory Regularization}
\label{subsubsec:gompertz_regularization}

To avoid sample-level parameter non-identifiability, the regularizer used a
single population-level Gompertz reference estimated from development data
only. Let $\bar{z}_{\mathrm{tr}}(t)$ denote the mean absolute log volume among
development observations at relative visit time $t$, with
$z_0=\bar{z}_{\mathrm{tr}}(t_0)$. The reference parameters were estimated by
nonlinear least squares:
\begin{equation}
(a_{\mathrm{tr}},b_{\mathrm{tr}})=
\underset{a>0,\,b}{\arg\min}
\sum_{t\in\mathcal{T}_{\mathrm{tr}}}
\left[
\bar{z}_{\mathrm{tr}}(t)-
\left\{
b-(b-z_0)\exp[-a(t-t_0)]
\right\}
\right]^2.
\label{eq:gomp_reference_fit}
\end{equation}

Duplicate observations at an identical relative time were averaged before the
fit. No calibration or test observation, outcome, or model prediction was used
to estimate the reference. During patient-grouped development
cross-validation, the reference parameters were re-estimated separately within
each training fold and then applied only to the corresponding validation
patients. After model selection, a final reference was estimated from the full
development cohort, yielding
$a_{\mathrm{tr}}=0.088805$ and
$b_{\mathrm{tr}}=1.649348$. These final values were then held fixed across
history lengths, training repeats, regularization weights, calibration
patients, and held-out test patients.

For sample $i$, let
$\Delta t_i=t_{i,\mathrm{target}}-t_{i,\mathrm{last}}$.
The discrete midpoint residual on the absolute log-volume scale was
\begin{equation}
r_i=
\frac{\hat{z}_i-z_{i,\mathrm{last}}}{\Delta t_i}
-a_{\mathrm{tr}}
\left(
b_{\mathrm{tr}}-
\frac{z_{i,\mathrm{last}}+\hat{z}_i}{2}
\right).
\label{eq:gomp_residual}
\end{equation}

The corresponding midpoint recurrence was
\begin{equation}
\tilde{z}_{i,\mathrm{target}}=
\frac{
\left(
1-\frac{a_{\mathrm{tr}}\Delta t_i}{2}
\right)
z_{i,\mathrm{last}}
+
a_{\mathrm{tr}}\Delta t_i b_{\mathrm{tr}}
}{
1+\frac{a_{\mathrm{tr}}\Delta t_i}{2}
}.
\label{eq:gomp_recurrence}
\end{equation}

This restriction is required for identifiability: if two trajectory-specific
parameters were free while only one transition equation was imposed, an
arbitrary prediction could be made to have zero residual. Estimating a shared
population reference from training data and holding it fixed removes this
degree of freedom.

The regularization loss and total objective were
\begin{equation}
\mathcal{L}_{g}
=
\frac{1}{N}
\sum_{i=1}^{N}
r_i^2,
\label{eq:gomp_loss}
\end{equation}
and
\begin{equation}
\mathcal{L}_{\mathrm{total}}
=
\mathcal{L}_{\mathrm{pred}}
+
\lambda\mathcal{L}_{g}.
\label{eq:gomp_total_objective}
\end{equation}

Residual reduction was interpreted as an expected consequence of the
regularized objective rather than evidence of Gompertz-like lesion dynamics.

\subsubsection{Statistical Inference}
\label{subsubsec:statistical_inference}

Statistical uncertainty was estimated using patient-level cluster bootstrap to
preserve dependence among multiple trajectories from the same patient.
Test-set patients were resampled with replacement, all corresponding
trajectories were retained, and metrics were recomputed for 5000 bootstrap
replicates. Percentile 95\% confidence intervals were
\begin{equation}
\mathrm{CI}_{95\%}(M)
=
\left[
Q_{0.025}(M^{*}),
Q_{0.975}(M^{*})
\right],
\label{eq:bootstrap_ci}
\end{equation}
where $M^{*}$ denotes the metric computed on a bootstrap-resampled test set.

RMSE, MAE, PICP, MPIW, interval score, WIS, and the imposed residual were
recomputed within each bootstrap sample. Raw-distribution PIT calibration error and NLL were bootstrapped only where a
predictive CDF and density were defined. Separately, variation
from stochastic model fitting was summarized as the mean, sample standard
deviation, and range of the ten seed-specific test metrics. Every seed was
evaluated on the same 38 trajectories from the same 26 held-out patients; seed
repeats were therefore not pooled into 380 independent test observations.
Bootstrap intervals reflect finite-sample uncertainty across held-out patients,
whereas the seed summaries describe training variability.

\subsubsection{Error--Uncertainty Association}
\label{subsubsec:error_uncertainty_association}

Sample-level uncertainty informativeness was assessed using Spearman's rank
correlation between calibrated prediction-interval width and absolute
prediction error. For sample $i$, the calibrated interval width was

\begin{equation}
w_i
=
\hat{u}_i-\hat{\ell}_i,
\label{eq:calibrated_interval_width}
\end{equation}

and the uncertainty--error association was

\begin{equation}
\rho_{w,e}
=
\operatorname{Spearman}
\left(
w_i,\,
|\hat{y}_i-y_i|
\right).
\label{eq:error_uncertainty_corr}
\end{equation}

A positive $\rho_{w,e}$ indicates that samples with larger prediction errors
tend to receive wider prediction intervals.


\section{Experimental Setup}
\label{sec:experimental_setup}


\subsection{Experimental Framework}
\label{subsec:experimental_framework}

All predictive analyses used the five-visit DeepLesion-DLT cohort described
in Sections~2.1--2.4 and predicted $T_4$ from subsets of the preceding
observations. Experiment~1 additionally included the exploratory conventional
sparse-to-final analysis defined in Section~2.6, whereas the primary
fixed visit-index horizon analyses used the most recent $m\in\{1,2,3,4\}$ pre-target
visits ending at $T_3$.

\begin{figure}[pos=h]
    \centering
    \includegraphics[width=0.6\linewidth]{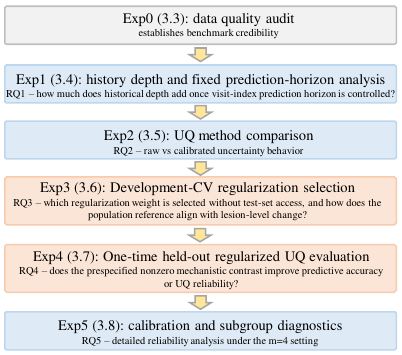}
    \caption{Overview of the six-part experimental framework. Experiment~0
    audits benchmark quality, followed by analyses of history depth, uncertainty quantification, Gompertz-inspired regularization, and calibration and subgroup reliability.}
    \label{fig:experimental_framework}
\end{figure}


\subsection{Common Experimental Settings}
\label{subsec:common_settings}

All experiments used the patient-level development/calibration/test split
defined in Section~\ref{subsec:data_split}. Model fitting, hyperparameter
selection, and regularization-setting selection were confined to the
development patients, using patient-grouped cross-validation where selection
was required. Calibration patients were used only for patient-level conformal
calibration, and test patients were reserved exclusively for final evaluation.

Models were trained independently within each experiment under the settings
reported in Tables~\ref{tab:implementation_settings} and
\ref{tab:experiment_training_settings}. For stochastic neural models, small differences in baseline
performance across experiments may therefore arise from random seeds,
repeat sets, and experiment-specific training protocols; comparisons are
interpreted relative to the matched baseline within each experiment. The Per-Trajectory Temporal GP used in Experiment 1 and the Cohort-Level Feature GP used in Experiment 2 were intentionally different baselines, and their RMSE values are therefore not interchangeable.

Experiment~1 used a fixed deterministic MLP across all history lengths. In the
UQ analyses, raw predictive distributions and patient-level conformal intervals
were evaluated separately using the metric domains defined in
Table~\ref{tab:evaluation_metrics}.

\begin{figure}[pos=h]
    \centering
    \includegraphics[width=0.5\linewidth]{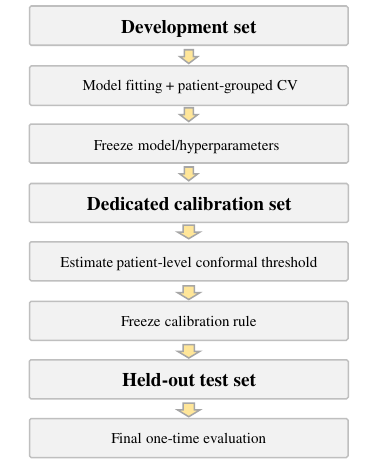}
    \caption{Development, calibration, and held-out test workflow. Model fitting
    and patient-grouped cross-validation are restricted to development patients;
    conformal thresholds use dedicated calibration patients. Test patients are
    evaluated only after model-fitting choices and complete sensitivity grids
    are fixed. Sensitivity experiments report every prespecified setting and do
    not use test results for hyperparameter selection.}
    \label{fig:evaluation_workflow}
\end{figure}

Figure~\ref{fig:evaluation_workflow} summarizes the evaluation workflow.

\subsection{Experiment 0: Data Quality Audit}
\label{subsec:exp0}

Experiment~0 assessed the integrity and composition of the constructed
DeepLesion-DLT benchmark before predictive modeling. The audit examined
candidate trajectory lengths, anatomical-site composition, patient-level split
counts, and outlier rates for raw volume, absolute log-volume, and
baseline-relative log-volume. These checks were used to verify the five-visit
trajectory construction and to compare alternative lesion-size representations
before defining the final forecasting outcome.


\subsection{Experiment 1: History Depth and Fixed Visit-index Prediction-Horizon Analysis}
\label{subsec:exp1}

Experiment~1 evaluated the two longitudinal prediction designs defined in
Section~\ref{subsec:analysis-designs}. The conventional sparse-to-final analysis
was treated as exploratory because history depth and the visit-index distance
between the most recent observation and $T_4$ varied simultaneously. The
fixed visit-index horizon recent-history analysis served as the primary
assessment of the incremental predictive value of older observations. Under
this design, the forecast endpoint was always $T_4$, the most recent observed
visit was always $T_3$, and the reported outcome was the common log change
$\Delta_{3\rightarrow4}$. Progressively earlier ordered observations were added
as $m$ increased. Here, $T_0$--$T_4$ denote ordered longitudinal observations
in the released DLT data rather than equally spaced calendar-time points. The
prediction model was fixed to the same deterministic MLP family for all $m$,
and the same model family was used in the pooled and anatomical-subgroup
analyses. This design reduced confounding by observation recency, visit-index
prediction horizon, and UQ-method selection.

Traditional longitudinal baselines were also evaluated under the same
patient-level split and prediction task. These included training-set mean and
median prediction, last observation carried forward, ridge trajectory
regression, linear extrapolation, Per-Trajectory Temporal GP
\cite{rasmussen2006gaussian}, and Gompertz/logistic curve fitting. Methods were
included only when they were identifiable from the available number of ordered
observations.

The Per-Trajectory Temporal GP used in Experiment~1 was a per-trajectory
temporal extrapolator. An RBF GP was fitted separately to the observed
visit-index--log-volume pairs of each trajectory. For test trajectories, the
GP was conditioned only on the pre-target observations available under the
corresponding history-length setting; the $T_4$ measurement was used only to
compute the final $T_3$-to-$T_4$ log-change outcome for performance evaluation.
The kernel length scale and noise level were selected using patient-grouped
cross-validation within the development cohort and were then fixed before
evaluation on the held-out test cohort. The GP did not learn cohort-level
regression coefficients across patients and did not use the fixed-length
feature representation employed by the cohort-level GP in Experiment~2.

Robustness to the five-visit trajectory definition was assessed using the
first five ordered observations, the latest five observations, sliding
five-observation windows, and randomly selected five-observation windows. All
derived windows retained the original patient-level split. RMSE and MAE were
the primary metrics for the follow-up-density, subgroup, baseline, and window
analyses.


\subsection{Experiment 2: UQ Method Comparison}
\label{subsec:exp2}

Experiment~2 compared Deterministic MLP, Cohort-Level Feature GP, MC Dropout
\cite{gal2016dropout}, Deep Ensemble
\cite{lakshminarayanan2017simple}, and the Gaussian residual-scale baseline under the
same prediction task and patient-level split.

Point-prediction performance was evaluated using RMSE and MAE of the
predictive mean. Raw probabilistic predictions were evaluated with PICP, MPIW, predictive-density NLL, and PIT calibration error. Patient-level
conformal intervals were evaluated at nominal 80\%, 90\%, and 95\% coverage
using trajectory-level PICP, MPIW, interval score, WIS, and the secondary
patient-level simultaneous coverage in
Eq.~\eqref{eq:patient_simultaneous_coverage}; calibrated NLL and PIT error
were not defined. Method selection used only patient-grouped cross-validation among
development patients.

\subsection{Experiment 3: Development-CV Regularization Selection and Reference Alignment}
\label{subsec:exp3}

Experiment~3 selected the regularization weight without accessing the held-out
test set. Within the development cohort, five patient-grouped folds were each
run with two training seeds, yielding ten fold--seed evaluations per history
length. For each fold, the Gompertz reference was estimated using only the
training patients in that fold, and
$\lambda\in\{0,0.1,1,10,100\}$ was evaluated on the corresponding validation
patients. The prespecified primary rule minimized development-CV RMSE at
$m=4$ over the full grid, with ties favoring the smaller weight. This procedure
selected $\lambda^*=0$. To retain a nonzero mechanistic comparison, the best
nonzero development-CV value, $\lambda=0.1$, was fixed separately before the
held-out test set was opened.

The same out-of-fold predictions were used to assess whether a population-level
reference aligned with lesion-level change. Observed $T_3\rightarrow T_4$
change was compared with the change implied by the fixed Gompertz reference
and with a generic linear mean-reversion control,
$\hat{y}_{T_4}=\beta_0+\rho y_{T_3}$. This comparison was used to distinguish
the general effect of shrinkage toward a population relationship from the
specific Gompertz functional form. RMSE, mean absolute imposed residual,
directional agreement, and growth-class-specific reference alignment were
reported from development out-of-fold predictions.

A separate prespecified measurement-proxy stress test retained $\lambda=1$ as
a moderate nonzero regularization setting and compared it with the
unregularized model at $m=4$. The analysis was repeated using ellipsoid volume
$(\pi/6)LS^2$, RECIST area $LS$, and long-axis length $L$. For each proxy, the
input was its absolute log value and the outcome was the corresponding
$T_3$-to-$T_4$ log change. This analysis was conducted only as a robustness
audit and was not used for hyperparameter selection or for defining the
held-out mechanistic contrast.


\subsection{Experiment 4: One-Time Held-Out Regularized UQ Evaluation}
\label{subsec:exp4}

Experiment~4 was the one-time held-out evaluation performed after the
Experiment~3 choices were fixed. The primary development-CV choice was
$\lambda^*=0$; for a prespecified mechanistic contrast, the best nonzero
choice $\lambda=0.1$ was compared with a matched unregularized MC Dropout
model. No test-set outcome was used to select or revise $\lambda$.

For each $m$, both models were trained for 300 epochs with ten random seeds.
MC Dropout remained active for 50 stochastic forward passes. Raw intervals
were constructed from the stochastic predictive mixture and patient-level
conformal calibration was then applied using the procedure in
Section~\ref{subsec:uq_calibration}. The 38 held-out trajectories from 26
patients were evaluated only after model and calibration choices had been
fixed.

Calibrated performance was assessed using RMSE, PICP, MPIW, interval score,
one-level WIS, and mean absolute Gompertz-style residual. NLL and PIT
calibration error were reported only for raw predictive distributions and were
not inferred from conformal interval widths. Because the residual appears
directly in the training objective, its reduction was used only to quantify
agreement with the fixed reference. The inferential question was whether the
nonzero development-selected contrast improved predictive accuracy or UQ
reliability relative to the matched unregularized model.


\subsection{Experiment 5: Calibration and Subgroup Reliability Diagnostics}
\label{subsec:exp5}

Experiment~5 examined the $m=4$ calibrated predictions from Experiment~2 using
pooled-cohort and subgroup reliability diagnostics.

At the pooled-cohort level, raw predictive PIT reliability diagrams,
calibrated interval-width distributions, interval width versus absolute
prediction error, and coverage stratified by target magnitude were examined. These analyses complemented PICP
by assessing interval sharpness and whether greater predictive uncertainty was
associated with larger observed errors \cite{gneiting2007strictly}.
Uncertainty--error association was summarized using Spearman's rank correlation
between absolute prediction error and calibrated interval width.

Subgroup analyses were performed separately for chest/lung, abdomen/liver,
and other lesions in one independent test evaluation. Coverage was reported as an integer numerator and denominator, with exact
Clopper--Pearson 95\% confidence intervals calculated at the trajectory level.
Because these intervals do not account for potential within-patient dependence
among multiple trajectories, they were used only as descriptive measures of
uncertainty for the exploratory subgroup analyses. MPIW and MAE were also
reported. All anatomical subgroup analyses were prespecified as exploratory
because of the small number of independent test patients and the heterogeneous
lesion composition. No confirmatory subgroup hypothesis testing was performed.
PIT calibration error and NLL were not inferred from conformal interval widths.


\subsection{Statistical Reporting and Confidence Intervals}
\label{subsec:statistical_reporting}

Statistical uncertainty was quantified using the patient-level cluster
bootstrap procedure defined in
Section~\ref{subsubsec:statistical_inference}. Main-text 95\% confidence
intervals were based on 5000 patient-level bootstrap replicates, preserving
dependence among multiple trajectories from the same patient.

Repeated random-seed runs were used only to assess training stability and
were not interpreted as independent test samples or generalization confidence
intervals. The held-out test cohort consisted of 38 trajectories from 26
patients; repeated evaluations across training seeds did not increase the
number of independent held-out test units. Coverage was reported as an integer
numerator and denominator for each run (for example, 37/38). When runs were
pooled descriptively, the denominator was explicitly labelled as
trajectory--model-repeat evaluations rather than unique test trajectories. For matched method comparisons, paired patient-level bootstrap was used so
that both methods were evaluated on the same resampled patients. 


\section{Results}
\label{sec:results}

\subsection{Experiment 0: Data Quality Audit of the DeepLesion-DLT Cohort}
\label{subsec:results_exp0}

The component-level audit began with 2,117 DLT connected components, all of which were patient-consistent. Repeated-pair deduplication affected all 2,117 components and collapsed 3,888 redundant pair annotations. A separate directionality audit confirmed that the 7,776 directed DLT records comprised 3,888 reciprocal pairs, split evenly between ascending and descending StudyID/ScanID orientation; source/target labels therefore did not define chronological direction. No components contained branching nodes, multiple same-study candidate nodes, or graph/study ambiguity, and none were excluded by the strict ambiguity rule. Requiring at least five unique CT visits retained 205 components, yielding a final benchmark of 205 trajectories from 129 patients. \par
\begin{figure}[pos=h]
    \centering
    \includegraphics[width=0.95\linewidth]{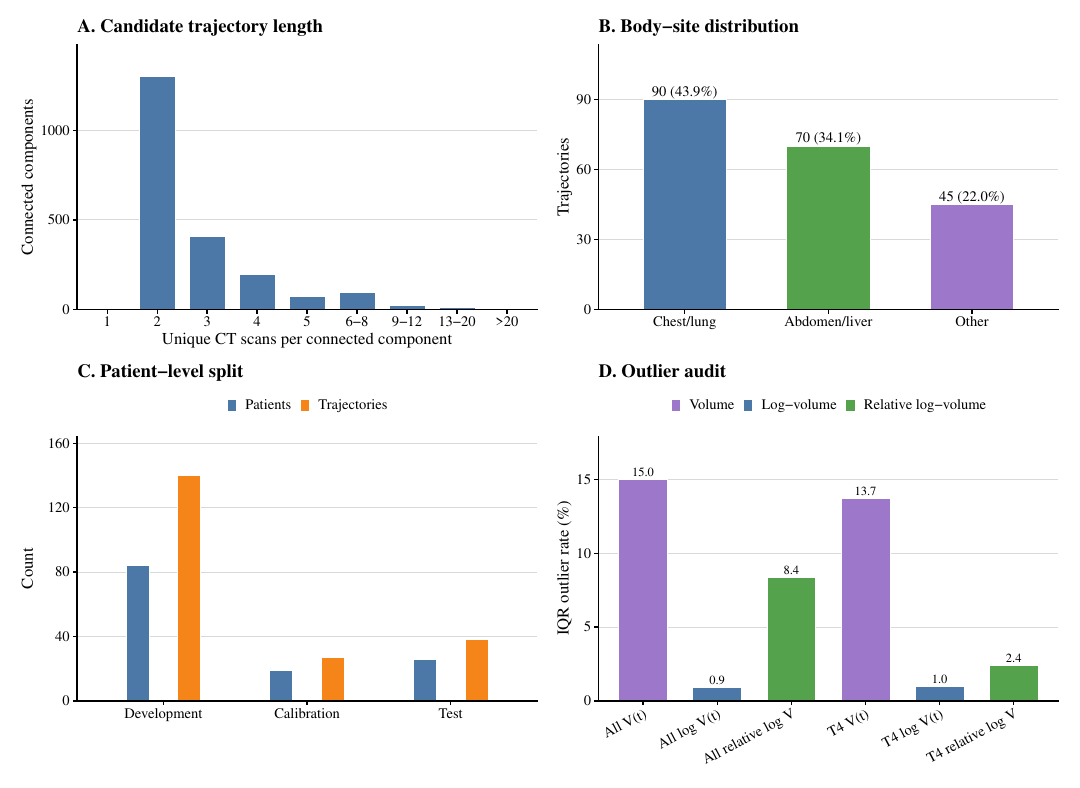}
    \caption{Data quality audit of the DeepLesion-DLT cohort
    (Experiment~0). Panels show candidate trajectory lengths (A),
    anatomical-site composition (B), patient-level data split (C), and
    IQR-based outlier rates for lesion-size variables (D).}
    \label{fig:data_quality_audit}
\end{figure}

 Figure~\ref{fig:data_quality_audit} summarizes the quality audit of the
constructed DeepLesion-DLT cohort. Although most DLT connected components
contained only two CT scans, 205 same-lesion trajectories had at least five
visits and were retained for the main benchmark.

The final cohort included 90 chest/lung trajectories (43.9\%), 70
abdomen/liver trajectories (34.1\%), and 45 trajectories from other sites
(22.0\%), confirming that the benchmark represents a mixed-lesion population.
The patient-level split contained 140 development, 27 calibration, and 38 test
trajectories without patient overlap.

The outlier audit showed that log transformation reduced the influence of
extreme lesion sizes, with absolute log volume exhibiting substantially fewer outliers than raw
volume. These findings supported log transformation before constructing the
primary $T_3$-to-$T_4$ change target.

\subsection{Experiment 1: History Depth and Fixed Visit-index Prediction-Horizon Analysis}
\label{subsec:results_exp1}
The exploratory conventional sparse-to-final analysis showed a substantially
larger apparent improvement as additional follow-up observations became
available. Because increasing $m$ simultaneously increased history depth and
moved the latest observation closer to $T_4$, these differences were interpreted
descriptively rather than as the isolated effect of additional history.

Under the primary fixed visit-index horizon design, all settings predicted
$\Delta_{3\rightarrow4}$ and all histories ended at $T_3$. For the fixed
deterministic MLP, held-out RMSE was 0.4726 at $m=1$, 0.4305 at $m=2$,
0.4499 at $m=3$, and 0.4513 at $m=4$. RMSE was lowest at $m=2$ and did not improve at $m=3$ or $m=4$;
adding the earliest observation $T_0$ did not improve the mean
result.

\begin{figure}[pos=h]
    \centering
    \includegraphics[width=0.9\linewidth]{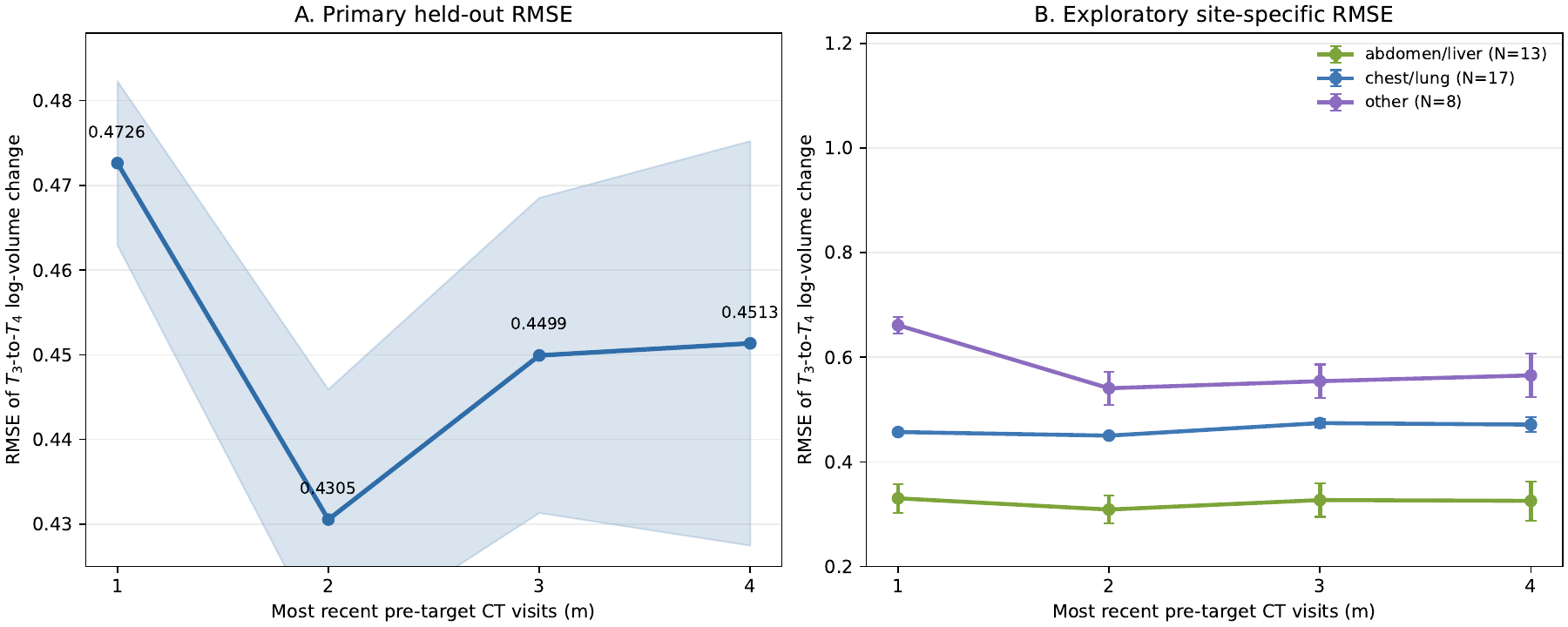}
    \caption{Fixed visit-index horizon recent-history experiment
    (Experiment~1). Every setting predicts
    $\Delta_{3\rightarrow4}=\log\{V(T_4)/V(T_3)\}$; $m=1,2,3,4$ use absolute
    log-volume histories at $T_3$, $(T_2,T_3)$, $(T_1,T_2,T_3)$, and
    $(T_0,T_1,T_2,T_3)$, respectively. Panel A reports mean held-out RMSE
    across ten training seeds; shading summarizes training variability across
    repeated fits. Panel B provides an exploratory site-specific analysis from
    ten repeated deterministic fits; error bars summarize training variability
    across those fits.}
    \label{fig:followup_density}
\end{figure}

Figure~\ref{fig:followup_density} reports the same model in the pooled test set
and anatomical subgroups. The subgroup curves were not uniformly monotonic,
particularly for the eight heterogeneous \textit{other} trajectories. The
usefulness of remote history may therefore depend on lesion site and should not
be inferred from the pooled curve alone.

Traditional baselines remained competitive under the same fixed target and
recent-history windows. At $m=1$, last observation carried forward achieved
RMSE 0.4530 and ridge trajectory regression achieved 0.4578, compared with
0.4726 for the deterministic MLP. At $m=4$, least-squares linear extrapolation
achieved RMSE 0.4191, ridge regression 0.4356, and the deterministic MLP 0.4513.
The Per-Trajectory Temporal GP had RMSE 0.6501 at $m=4$. This per-trajectory
extrapolator is distinct from the Cohort-Level Feature GP evaluated in
Experiment~2; the two GP results are therefore not repeated estimates of the
same model.

Overall, once $T_3$, the $T_3\rightarrow T_4$ outcome, and the model family
were fixed, progressively older observations provided only limited incremental
predictive benefit.

\subsection{Experiment 2: UQ Method Comparison}
\label{subsec:results_exp2}

Experiment~2 compared Deterministic MLP, Cohort-Level Feature GP, MC Dropout,
Deep Ensemble, and the Gaussian residual-scale baseline under matched
conditions. At $m=4$, mean RMSE across training repeats was 0.4513, 0.5170,
0.4741, 0.4266, and 0.4513, respectively. Deep Ensemble had the lowest mean
error. As required by its construction, the Gaussian residual-scale baseline
exactly matched the Deterministic MLP point predictions and RMSE in every
repeat; it differed only by estimating a single development residual scale.

Each of the ten seeds was evaluated on the same 38 held-out trajectories from
26 patients. Raw 95\% PICP at $m=4$ was 0.9474 for Cohort-Level Feature GP,
1.0000 for MC Dropout, and 0.9974 for both Deep Ensemble and Gaussian
residual-scale. Corresponding raw MPIW was 2.4485, 2.8788, 2.2976, and 2.4405.
The PIT calibration errors were 0.0731, 0.1134, 0.0710, and 0.0981, and
mixture-consistent NLL was 0.7988, 0.8052, 0.6478, and 0.7089, respectively.
Thus, the two stochastic neural methods were evaluated using their full
Gaussian-mixture CDF and density rather than a Gaussian scale reconstructed
from interval width. Seed summaries describe training variability and do not
increase the number of independent test observations.

At nominal 95\% coverage, the patient-level conformal rank was 19 of 19, so
the adjustment used the maximum calibration-patient score. Calibrated
trajectory-level PICP at $m=4$ was 1.0000 for Deterministic MLP, 0.9474 for
Cohort-Level Feature GP, and 1.0000 for MC Dropout, Deep Ensemble, and Gaussian
residual-scale; patient-level simultaneous coverage was 1.0000, 0.9231,
1.0000, 1.0000, and 1.0000, respectively. Calibrated MPIW was 3.0783, 2.5978,
3.2776, 2.7889, and 3.0783. These near-nominal or conservative,
resolution-limited results were not treated as evidence of method superiority.
Conformalized intervals do not define a predictive density or CDF, so
calibrated NLL and PIT calibration error were not reported.

\begin{figure}[pos=htbp]
    \centering
    \includegraphics[width=0.95\linewidth]{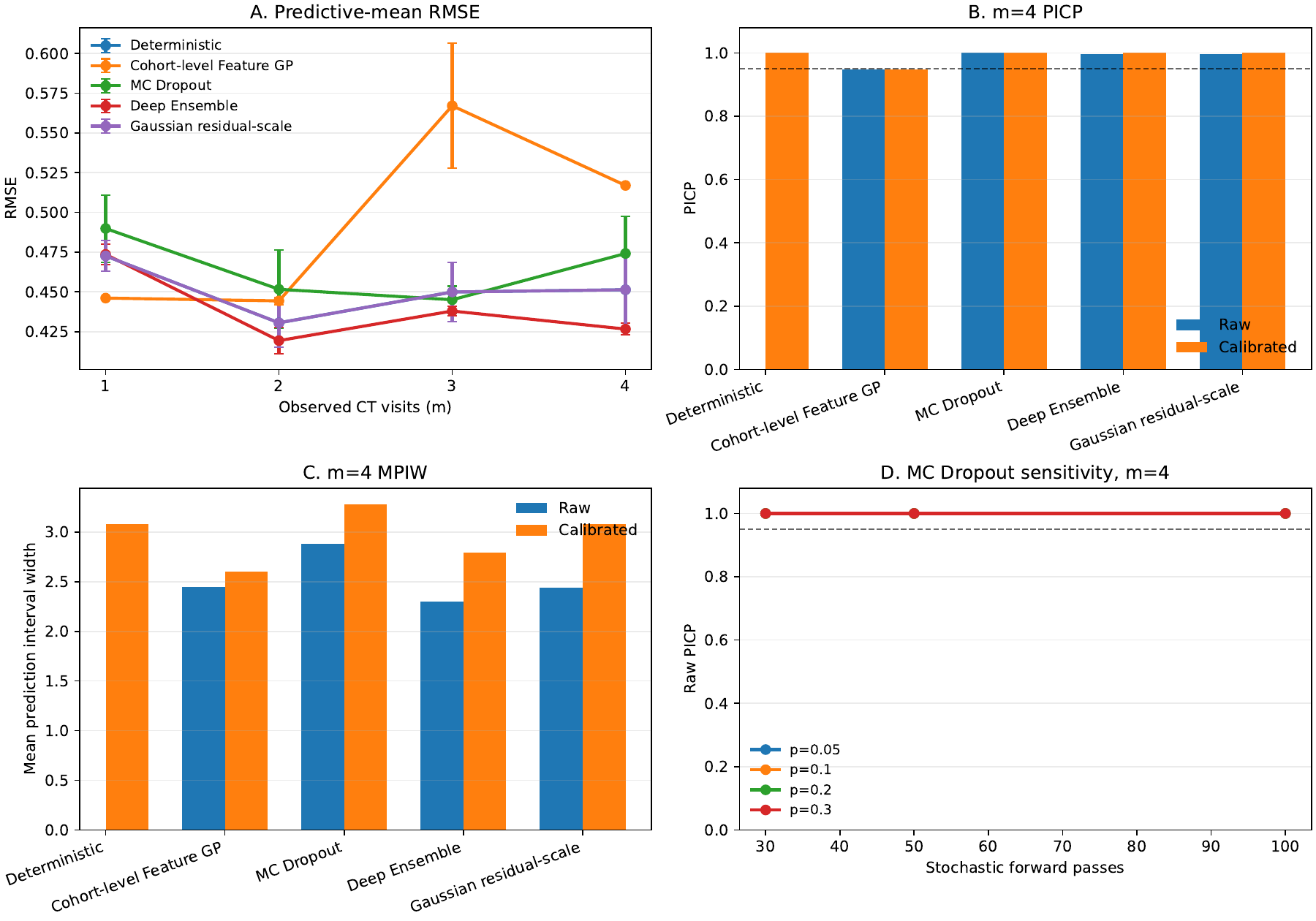}
    \caption{Uncertainty-method comparison for the common
    $T_3\rightarrow T_4$ log-change task (Experiment~2). Panel A shows
    predictive-mean RMSE across recent-history lengths. Panels B and C compare
    raw and patient-level conformal coverage and interval width at $m=4$.
    Panel D reports MC Dropout sensitivity. Each seed was evaluated on the same
    38 trajectories from 26 patients; summaries describe training variability
    and do not treat repeated evaluations as independent test samples.}
    \label{fig:uq_comparison}
\end{figure}

\begin{figure}[pos=htbp]
    \centering
    \includegraphics[width=0.9\linewidth]{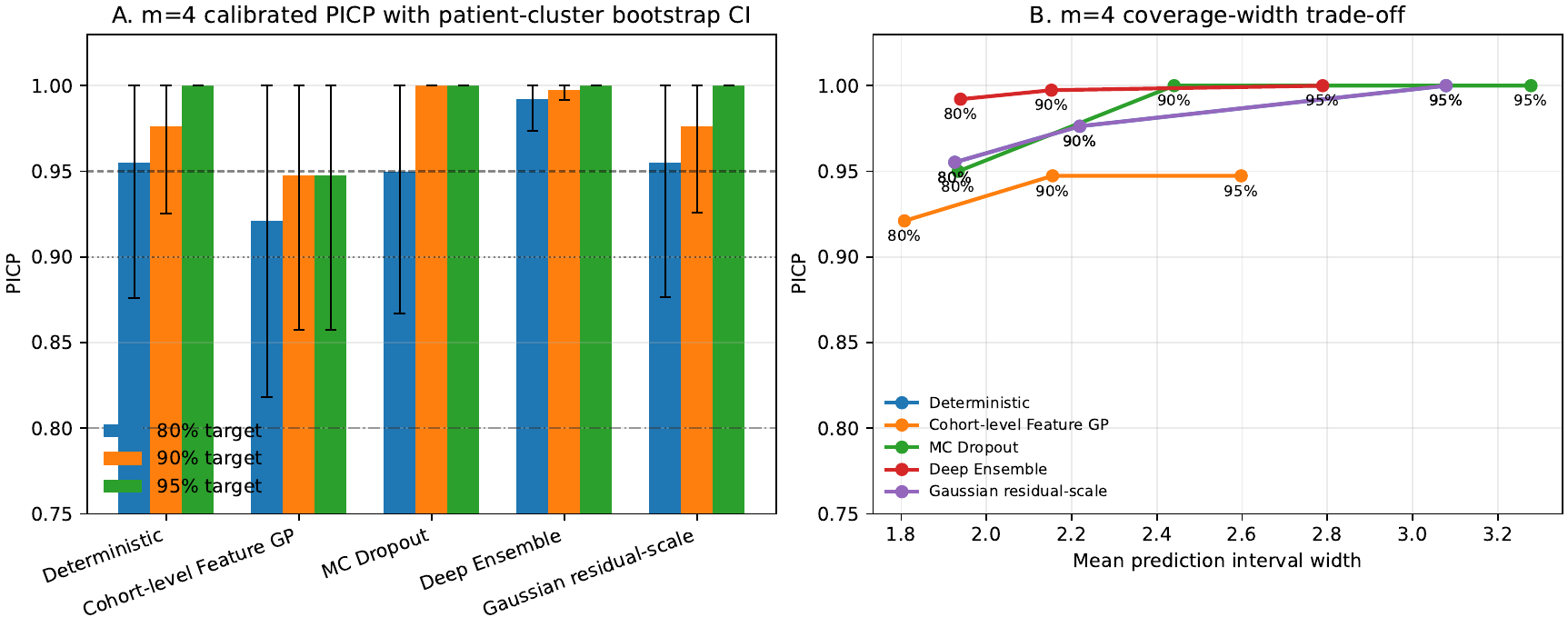}
    \caption{Calibration analysis at 80\%, 90\%, and 95\% nominal coverage
    for $m=4$. Panel A shows trajectory-level PICP with 95\% patient-cluster
    bootstrap intervals from 5,000 resamples of 26 test patients (38
    trajectories); Panel B shows the corresponding coverage--width trade-off.
    Interval score and patient-level simultaneous coverage are secondary
    endpoints.}
    \label{fig:calibration_sensitivity}
\end{figure}

The 80\%, 90\%, and 95\% analyses exposed the coverage--sharpness trade-off
rather than presenting complete 95\% coverage as an advantage. For $m=4$ MC
Dropout, trajectory-level PICP was 0.9500, 1.0000, and 1.0000; MPIW was 1.9339,
2.4404, and 3.2776; and interval score was 1.9902, 2.4404, and 3.2776,
respectively. Patient-level simultaneous coverage was 0.9269, 1.0000, and
1.0000. Figure~\ref{fig:calibration_sensitivity} uses patient-level cluster
bootstrap intervals so that all trajectories from a resampled patient remain
together.

Overall, near-nominal or conservative marginal coverage did not imply equally
sharp intervals, well-calibrated predictive distributions, or equally accurate
point predictions. These quantities therefore require separate interpretation.


\subsection{Experiment 3: Development-CV Regularization Selection and Reference Alignment}
\label{subsec:results_exp3}

Experiment~3 confined the complete $\lambda$ grid to patient-grouped
development cross-validation. For $m=4$, out-of-fold RMSE was 0.6385
$\pm$ 0.0560 at $\lambda=0$, and 0.8993 $\pm$ 0.2527, 1.2388
$\pm$ 0.3339, 1.3982 $\pm$ 0.3468, and 1.4188 $\pm$ 0.3489 at
$\lambda=0.1,1,10,100$, respectively. The prespecified full-grid rule
therefore selected $\lambda^*=0$. Conditional on carrying a nonzero
mechanistic contrast forward, $\lambda=0.1$ was the best nonzero choice.
Both choices were fixed before the held-out test set was opened.

Agreement with the frozen reference increased as expected with larger weights.
At $m=4$, the mean absolute imposed residual was 3.2117 without regularization
and 1.2442, 0.3269, 0.0612, and 0.0334 at increasing nonzero weights. Because
this residual appears directly in the training objective, its reduction is an
optimization effect and not independent evidence of biological validity.

The out-of-fold alignment diagnostic clarified the mechanism. The fixed
Gompertz population reference agreed in sign with only 45.7\% of observed
lesion-level $T_3\rightarrow T_4$ changes; a generic linear mean-reversion
control agreed in 48.6\%. At $m=4$, the generic control had RMSE 0.6449,
close to the unregularized MLP development-CV RMSE of 0.6385 and far below the
nonzero Gompertz-regularized values, but it did not improve prediction. The
reference implied mean increases for decreasing lesions ($+0.413$ versus an
observed $-0.255$) and mean decreases for rapid-growth lesions ($-0.180$
versus an observed $+0.252$). Thus, a single population shrinkage target can
oppose heterogeneous lesion-level directions, explaining the observed
accuracy--consistency trade-off.

The separate measurement-proxy audit showed the same qualitative trade-off for
ellipsoid volume, long-axis length, and RECIST area, although its magnitude was
proxy-dependent. It was not used for hyperparameter selection.

\begin{figure}[pos=htbp]
    \centering
    \includegraphics[width=0.95\linewidth]{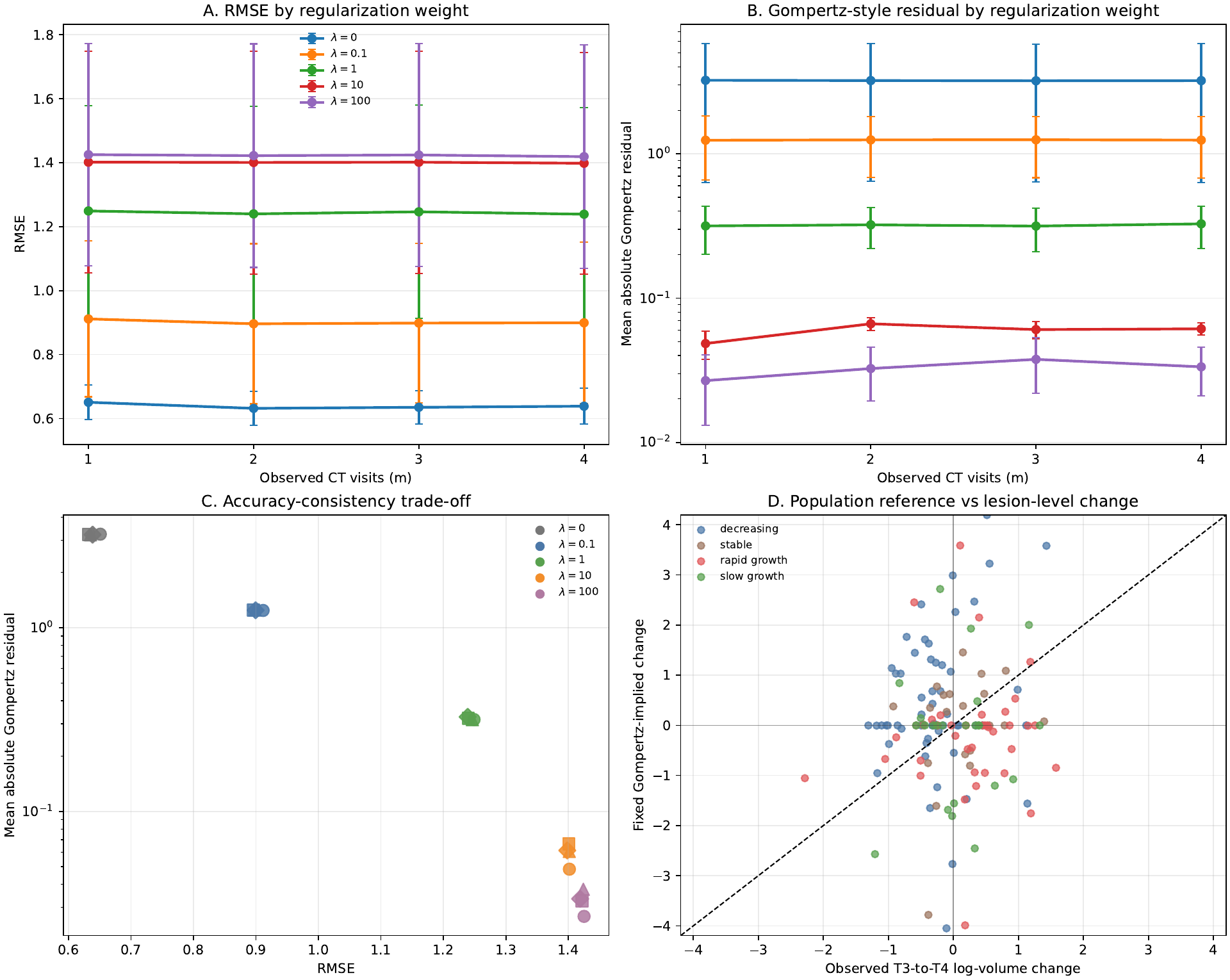}
    \caption{Development-only, patient-grouped cross-validation of the fixed
    Gompertz reference (Experiment~3). Panels show RMSE across regularization
    weights (A), mean absolute imposed residual on a logarithmic scale (B), the
    accuracy--consistency trade-off (C), and observed lesion-level change versus
    the change implied by the fixed population reference (D). Error bars
    summarize variability across the ten fold--seed evaluations. In Panel D,
    points with observed and reference-implied changes of opposite signs
    indicate lesions for which the population reference predicts the wrong
    direction of change. The held-out test set was not read; residual reduction
    measures agreement with the imposed reference rather than biological
    validity.}
    \label{fig:regularization_sensitivity}
\end{figure}


\subsection{Experiment 4: One-Time Held-Out Regularized UQ Evaluation}
\label{subsec:results_exp4}

Experiment~4 opened the held-out test set once after development-CV selection.
The primary choice was $\lambda^*=0$; the best nonzero development choice,
$\lambda=0.1$, was retained only as a mechanistic contrast against matched
unregularized MC Dropout UQ. Across $m=1,2,3,4$, the nonzero contrast improved
RMSE at 1/4 history lengths and interval score and one-level WIS at 2/4 each.
Its mean residual ratio relative to the unregularized model was 1.008, so it did
not consistently increase even the imposed reference agreement on test data.

Calibrated 95\% coverage was 1.0000 for both models because the patient-level
conformal rank used the largest of 19 calibration-patient scores. At $m=4$,
the unregularized model had RMSE 0.4557 $\pm$ 0.0177, interval score 3.3817
$\pm$ 0.2017, WIS 0.5128 $\pm$ 0.0207, and residual 0.1542 $\pm$
0.0460. The $\lambda=0.1$ model yielded 0.4607 $\pm$ 0.0391, 3.3038
$\pm$ 0.2635, 0.5098 $\pm$ 0.0346, and 0.1844 $\pm$ 0.0711,
respectively. These mixed, overlapping results did not support a uniform
accuracy or reliability benefit, and they were not used to revise the
preselected $\lambda^*=0$ conclusion.

\begin{figure}[pos=htbp]
    \centering
    \includegraphics[width=0.95\linewidth]{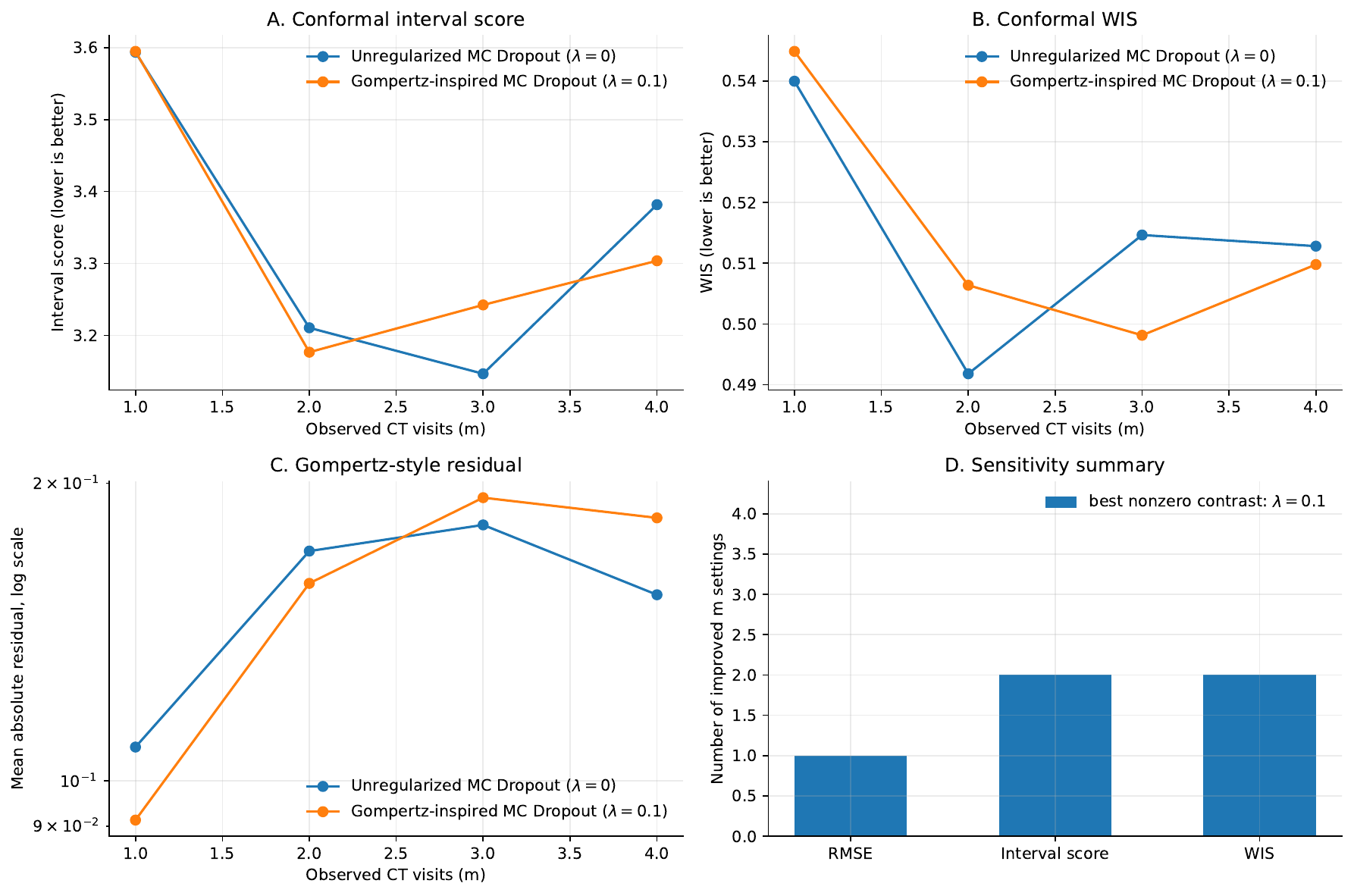}
    \caption{One-time held-out evaluation of the best nonzero
    development-CV contrast, $\lambda=0.1$, against matched unregularized MC
    Dropout UQ (Experiment~4; ten seeds). Panels compare calibrated interval
    score (A), one-level WIS (B), mean absolute imposed residual (C), and the
    number of history lengths with lower RMSE, interval score, or WIS (D). The
    primary development-CV choice was $\lambda^*=0$. Conformal NLL and PIT
    calibration error are not defined.}
    \label{fig:reg_uq_reliability}
\end{figure}


\subsection{Experiment 5: Exploratory Subgroup Reliability Diagnostics}
\label{subsec:results_exp5}

Experiment~5 examined pooled and subgroup reliability at $m=4$. Calibrated
PICP was 1.0000 for Deterministic MLP, MC Dropout, Deep Ensemble, and Gaussian
residual-scale, and 0.9474 for Cohort-Level Feature GP. Calibrated MPIW was
3.0783, 3.2776, 2.7889, 3.0783, and 2.5978 for Deterministic MLP, MC Dropout,
Deep Ensemble, Gaussian residual-scale, and Cohort-Level Feature GP,
respectively. The equality between Deterministic MLP and Gaussian
residual-scale reflects reuse of the same point predictor and the same
patient-level conformal residual calibration.

Distributional diagnostics used the raw predictive CDF rather than
conformalized interval width. PIT calibration error was 0.0731 for
Cohort-Level Feature GP, 0.1134 for MC Dropout, 0.0710 for Deep Ensemble, and
0.0981 for Gaussian residual-scale. For the mixture methods, these values used
the equal-weight Gaussian-mixture CDF. The association between calibrated
interval width and absolute error was weak for Cohort-Level Feature GP and Deep
Ensemble ($\rho_{w,e}=0.0004$ and 0.0391) and more variable for MC Dropout
(mean $\rho_{w,e}=0.2193$ across seeds). It was undefined for Deterministic
MLP and Gaussian residual-scale because their calibrated widths had no
meaningful sample-level variation.

Exploratory subgroup results were imprecise. Cohort-Level Feature GP covered
16/17 chest/lung trajectories, 13/13 abdomen/liver trajectories, and 7/8
\textit{other} trajectories; corresponding exact 95\% Clopper--Pearson
intervals were 0.7131--0.9985, 0.7529--1.0000, and 0.4735--0.9968. Other
methods showed complete subgroup coverage with wider intervals. These
trajectory-level intervals were interpreted descriptively because of small
subgroups and possible within-patient dependence.

Target-magnitude stratification showed the same limitation. For Cohort-Level
Feature GP, middle-change coverage was 14/14, while low- and high-change
coverage was 11/12 in each group. Other methods achieved complete coverage,
but each stratum contained only 12--13 trajectories.

\begin{figure}[pos=htbp]
    \centering
    \includegraphics[width=0.95\textwidth]{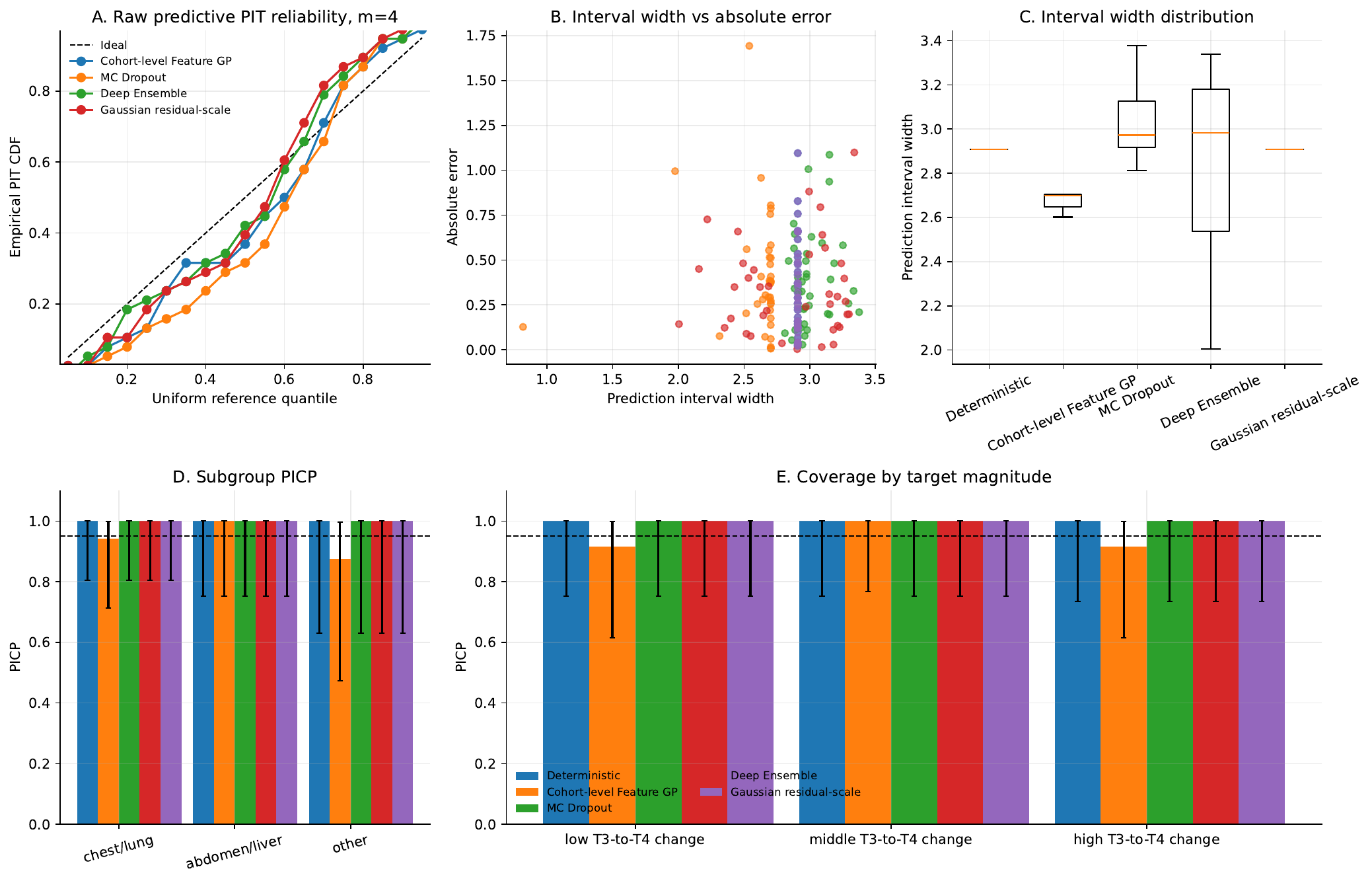}
    \caption{Calibration and subgroup reliability diagnostics for the
    $T_3\rightarrow T_4$ log-change outcome at $m=4$ (Experiment~5). Panel A
    shows the empirical CDF of raw predictive PIT values against the uniform
    reference; mixture methods use their full Gaussian-mixture CDF. Panels B
    and C show calibrated interval width versus absolute error and interval
    width distributions; Panels D and E show subgroup and target-magnitude
    coverage with exact Clopper--Pearson 95\% confidence intervals.}
    \label{fig:calibration_diagnostics}
\end{figure}

Overall, high marginal coverage coexisted with differences in sharpness, probabilistic calibration, case-level uncertainty--error alignment, and subgroup prediction difficulty.


\section{Discussion}
\label{sec:discussion}

This study evaluated sparse longitudinal CT lesion-size forecasting from a
reliability-centered perspective using an automatically curated five-visit
same-lesion DeepLesion-DLT benchmark. Three findings were consistent across
the analyses. First, when the target visit, most recent observation, and
visit-index prediction horizon were held fixed, adding one additional recent
observation improved prediction, whereas progressively earlier history beyond
$m=2$ provided little additional benefit; conventional longitudinal models
also remained competitive. Second, similar point-prediction accuracy did not
imply similar uncertainty behavior, and post-hoc calibration improved marginal
coverage without necessarily producing informative sample-level uncertainty.
Third, stronger Gompertz-inspired regularization increased agreement with the
imposed trajectory reference but did not translate into more accurate or more
reliable forecasts. Together, these results show that predictive accuracy,
uncertainty reliability, and trajectory consistency capture related but
distinct aspects of model performance.

\subsection{Historical Information and Prediction Accuracy}
\label{subsec:discussion_prediction}

The fixed visit-index horizon analysis was designed to separate the value of
additional history from the effect of moving the most recent observation
closer to the prediction target. Under this design, the improvement was
concentrated in the transition from $m=1$ to $m=2$; adding progressively
earlier observations provided no further reduction in prediction error. This
suggests that, in the present cohort, recent lesion history carried more useful
forecasting information than increasingly remote observations.

Traditional baselines further qualified this finding. Linear extrapolation and
ridge regression remained competitive with the deterministic MLP, whereas the
Per-Trajectory Temporal GP performed less well under the same fixed-horizon
task. Increasing model complexity therefore did not consistently improve point
prediction, emphasizing the importance of controlled prediction targets,
appropriate history windows, and strong conventional baselines.

This interpretation complements longitudinal deep-learning studies that
forecast future lesion appearance or volume
\cite{tao2022prediction,zhang2020spatiotemporal}. The present study addresses
a narrower question: whether progressively earlier observations add predictive
value when the current lesion scale, most recent visit, and prediction horizon
are held fixed. Under this setting, follow-up count alone did not correspond
directly to additional predictive information.

\subsection{Uncertainty Reliability and Trajectory Regularization}
\label{subsec:discussion_uq_physics}

Point-prediction accuracy, probabilistic calibration, and interval coverage
provided different views of model performance. The evaluated methods produced
different raw interval widths and calibration behavior even when their point
errors were similar. For MC Dropout and Deep Ensemble, probabilistic
calibration was evaluated using the full Gaussian-mixture predictive CDF and
density rather than a Gaussian approximation derived from interval width.
This distinction is important because interval coverage alone does not
characterize the quality of the underlying predictive distribution.

The Gaussian residual-scale baseline provided a useful control because it
shared exactly the same point predictor as the Deterministic MLP and differed
only in its uncertainty construction. Any difference between these two methods
therefore reflects interval estimation rather than a change in the predictive
mean.

Patient-level conformal calibration generally produced near-nominal or
conservative coverage but also widened intervals. At the 95\% nominal level,
the finite-sample rank was determined by the largest score among the 19
calibration patients, limiting the resolution of the upper-tail calibration.
Trajectory-level PICP should therefore be interpreted together with
patient-level simultaneous coverage, interval width, and proper interval
scores \cite{vovk2005algorithmic,lei2018distribution}. Because conformalized
intervals do not define a predictive density or CDF, NLL and PIT calibration
error were not assigned after calibration.

The regularization experiments further separated agreement with a structural
reference from forecasting performance. Development cross-validation selected
no Gompertz-inspired regularization, and both the fixed Gompertz reference and
the generic linear mean-reversion control showed limited directional agreement
with lesion-level change. In particular, a single population-level reference
could oppose the direction of change for decreasing or rapidly growing
lesions. The resulting trade-off therefore appears to reflect a broader
limitation of population-level shrinkage rather than a feature unique to the
Gompertz functional form.

The held-out comparison was consistent with this interpretation: the nonzero
regularized model did not show a uniform advantage in predictive accuracy,
interval quality, or agreement with the imposed reference. Future structural
priors may therefore benefit from being trajectory-conditioned or
mechanistically stratified, with their biological plausibility evaluated
separately from their ability to reduce an imposed training residual.

\subsection{Clinical Interpretation}
\label{subsec:discussion_limitations}

The subgroup analysis provides an exploratory view of how the findings vary
across anatomical categories. Improvement from $m=1$ to $m=2$ was also
observed for chest/lung and abdomen/liver lesions, whereas the smaller
``other'' subgroup showed less stable behavior. This is consistent with
broader concerns that aggregate performance can mask clinically relevant
subgroup variation in medical imaging \cite{oakdenrayner2020hidden}. These
subgroup results should therefore be interpreted as exploratory and
hypothesis-generating rather than as confirmatory evidence of anatomical
differences.

From a clinical perspective, additional longitudinal observations should not
be assumed to improve lesion-size forecasting simply because more scans are
available. In the fixed-horizon analysis, the benefit was concentrated in the
most recent additional observation, while older history provided little further
gain. Conventional longitudinal models also remained competitive with the
neural methods. The value of additional follow-up history may therefore depend
more on its relevance to an individual lesion trajectory than on the number
of available observations alone.

The uncertainty analyses likewise show that similar point-prediction accuracy
does not guarantee similar predictive reliability. Models with comparable
errors produced different interval widths, calibration behavior, and
uncertainty--error associations. Point accuracy should therefore be interpreted
together with coverage, sharpness, probabilistic calibration, and
sample-level uncertainty informativeness. Similarly, closer agreement with an
imposed trajectory reference should not by itself be interpreted as improved
forecasting reliability. These results support treating predictive accuracy,
uncertainty reliability, subgroup behavior, and trajectory consistency as
complementary rather than interchangeable dimensions of model performance.


\section{Limitations}
\label{sec:limitations}

Several limitations should be considered when interpreting these results.

First, DeepLesion-DLT is a heterogeneous mixed-lesion resource rather than a
disease-specific cohort. Requiring at least five identifiable same-lesion CT
visits yielded 205 trajectories and may preferentially select lesions with
longer follow-up or greater trackability. Although anatomical subgroup
analyses were performed, the resulting cohort should not be considered
representative of all DeepLesion lesions or broader clinical populations.

Second, lesion size was represented by a RECIST-based ellipsoid proxy rather
than three-dimensional segmentation volume
\cite{eisenhauer2009recist}. Two-dimensional measurements are subject to
measurement variability \cite{revel2004two}, so the resulting target should
be interpreted as a reproducible lesion-size proxy rather than exact tumor
volume. DLT-provided lesion matches were subjected to patient-, scan-, and
measurement-level quality control but were not manually re-verified by
radiologists. The automated audit identified 2,117 connected components,
3,888 repeated pair annotations, no unique-neighbor branching components, and
no multiple-candidate studies; no trajectory was selected using future lesion
size or trajectory smoothness. A 50-trajectory audit manifest was generated
for future image-level review, but no completed manual audit is claimed here.
Sensitivity analyses using ellipsoid volume, RECIST area, and long-axis length
showed the same qualitative regularization trade-off, although none substitutes
for three-dimensional segmentation.

Third, exact CT acquisition dates were not reliably available for all
trajectories. Longitudinal position was therefore represented by the
operational ordering derived from StudyID and ScanID, and
\(T_0,\ldots,T_4\) should be interpreted as ordered visit indices rather than
observations at verified chronological times or known elapsed clinical
intervals. Irregular follow-up intervals and possible differences between
identifier-based and true temporal ordering may therefore affect extrapolation
and the interpretation of the Gompertz-style residual. Although alternative
five-visit windows reproduced the general follow-up-density pattern, absolute
performance varied with window definition. Future studies should incorporate
verified acquisition dates, actual inter-scan intervals, and continuous-time
longitudinal models
\cite{rubanova2019latent,kidger2020neural,debrouwer2019gru}.

Fourth, treatment information and downstream clinical outcomes were
unavailable. Observed lesion-size changes may reflect progression, treatment
response, stabilization, or measurement variability, which cannot be
distinguished in the present data. The $T_3$-to-$T_4$ log-change target should
therefore be interpreted as a lesion-size forecasting endpoint rather than a
direct measure of prognosis or treatment response.

Finally, both the study cohort and the dedicated calibration set were modest
for deep-learning and uncertainty-quantification analyses. Although model
selection and calibration were separated at the patient level, only 19
calibration patients were available; the 95\% finite-sample conformal rank
therefore used the largest patient-level score and may produce conservative
intervals \cite{lei2018distribution}. Subgroup estimates were likewise based
on small samples. Larger independent cohorts are needed to assess
generalizability, subgroup reliability, and the clinical value of trajectory
regularization \cite{collins2015tripod}.

These limitations position the present work as a benchmark and reliability
analysis rather than a clinically validated forecasting system. External
validation should prioritize larger disease-specific cohorts with reliable
scan dates, treatment information, three-dimensional lesion measurements,
dedicated calibration data, and clinically meaningful endpoints.


\section{Conclusion}
\label{sec:conclusion}

This study established a five-visit DLT-derived longitudinal lesion benchmark
for reliability-centered evaluation of sparse longitudinal CT lesion-size
forecasting. Under a fixed $T_3\rightarrow T_4$ visit-index horizon, prediction
improved when one additional recent observation was included, but progressively
older history provided no further benefit, while conventional longitudinal
methods remained competitive. Models with similar point-prediction accuracy
showed different uncertainty behavior, and post-hoc conformal calibration
improved marginal coverage without necessarily providing informative
sample-level uncertainty.

Patient-grouped development cross-validation selected no Gompertz-inspired
regularization. The fixed population reference often conflicted with
lesion-level change directions, and the prespecified nonzero held-out contrast
did not consistently improve forecasting accuracy or uncertainty reliability.
Together, these findings show that predictive accuracy, uncertainty reliability,
and trajectory consistency should be evaluated as distinct but complementary
aspects of sparse longitudinal imaging prediction.


\section*{Declarations}

\subsection*{Ethics Approval}

This study is a secondary analysis of de-identified imaging-derived data from
the publicly available DeepLesion and DeepLesion Tracking (DLT) resources.
No new patient recruitment, intervention, or direct access to identifiable
patient information was involved. Therefore, no additional institutional
ethics approval was required for the present secondary analysis. All data were
used in accordance with the access and data-use conditions associated with the
original datasets.

\subsection*{Consent to Participate}

No new participants were recruited for this study. The analysis used only
previously collected, de-identified imaging-derived data, and no additional
participant contact or consent was required for the present secondary analysis.

\subsection*{Declaration of Competing Interests}

The author declares that there are no known competing financial interests
or personal relationships that could have appeared to influence the work
reported in this paper.

\subsection*{Funding}

This research received no specific grant from any funding agency in the public,
commercial, or not-for-profit sectors.

\subsection*{Data Availability}

This study used the publicly available DeepLesion and DeepLesion Tracking
(DLT) resources. The original third-party data are not redistributed.
Processing scripts, trajectory-construction code, data-format instructions,
and derived outputs required to reproduce the reported analyses are available
at:

\begin{quote}
\texttt{https://github.com/Lorettakong/deeplesion-dlt-reliability}
\end{quote}

The repository provides instructions for reconstructing the benchmark from
the original public releases.

\subsection*{Code Availability}

Code for data processing, trajectory construction, prediction modeling,
uncertainty quantification, calibration, statistical analysis, and figure
generation is publicly available at:

\begin{quote}
\texttt{https://github.com/Lorettakong/deeplesion-dlt-reliability}
\end{quote}

The repository includes environment requirements and instructions for
reproducing the reported experiments, tables, and figures.

\section*{CRediT authorship contribution statement}

\textbf{Lingfei Kong:} Conceptualization, Methodology, Software, Validation,
Formal analysis, Investigation, Data curation, Visualization,
Writing -- original draft, Writing -- review \& editing.

\subsection*{Acknowledgments}

The author acknowledges the support of the Department of Mathematics at
Vanderbilt University and the guidance of Prof. Glenn F. Webb during the
development of this research.


\bibliographystyle{model1-num-names}
\bibliography{cas-refs}

@article{cai2026pulmonary,
  author = {Cai, Ruichu and Zhao, Haifeng and Yan, Yuguang and He, Ke and Yan, Jianhao and Liu, Baichuan},
  title = {Pulmonary nodule growth prediction with anisotropic reaction--diffusion},
  journal = {Computer Methods and Programs in Biomedicine},
  volume = {284},
  pages = {109443},
  year = {2026},
  doi = {10.1016/j.cmpb.2026.109443}
}

@article{yan2018deeplesion,
  author = {Yan, Ke and Wang, Xiaosong and Lu, Le and Summers, Ronald M.},
  title = {{DeepLesion}: automated mining of large-scale lesion annotations and universal lesion detection with deep learning},
  journal = {Journal of Medical Imaging},
  volume = {5},
  number = {3},
  pages = {036501},
  year = {2018},
  doi = {10.1117/1.JMI.5.3.036501}
}

@inproceedings{cai2021deep,
  author = {Cai, Junyi and Tang, Youbao and Yan, Ke and Harrison, Adam P. and Xiao, Jing and Lin, Guosheng and Lu, Le},
  title = {Deep Lesion Tracker: monitoring lesions in {4D} longitudinal imaging studies},
  booktitle = {Proceedings of the IEEE/CVF Conference on Computer Vision and Pattern Recognition (CVPR)},
  pages = {15159--15169},
  year = {2021},
  doi = {10.1109/CVPR46437.2021.01491}
}

@inproceedings{gal2016dropout,
  author = {Gal, Yarin and Ghahramani, Zoubin},
  title = {Dropout as a Bayesian approximation: representing model uncertainty in deep learning},
  booktitle = {Proceedings of the 33rd International Conference on Machine Learning (ICML)},
  series = {Proceedings of Machine Learning Research},
  volume = {48},
  pages = {1050--1059},
  year = {2016}
}

@inproceedings{lakshminarayanan2017simple,
  author = {Lakshminarayanan, Balaji and Pritzel, Alexander and Blundell, Charles},
  title = {Simple and scalable predictive uncertainty estimation using deep ensembles},
  booktitle = {Advances in Neural Information Processing Systems},
  volume = {30},
  pages = {6402--6413},
  year = {2017}
}

@book{vovk2005algorithmic,
  author = {Vovk, Vladimir and Gammerman, Alexander and Shafer, Glenn},
  title = {Algorithmic Learning in a Random World},
  publisher = {Springer},
  address = {New York},
  year = {2005},
  doi = {10.1007/b106715}
}

@article{gompertz1825nature,
  author = {Gompertz, Benjamin},
  title = {On the nature of the function expressive of the law of human mortality, and on a new mode of determining the value of life contingencies},
  journal = {Philosophical Transactions of the Royal Society of London},
  volume = {115},
  pages = {513--583},
  year = {1825},
  doi = {10.1098/rstl.1825.0026}
}

@article{laird1964dynamics,
  author = {Laird, Anna K.},
  title = {Dynamics of tumor growth},
  journal = {British Journal of Cancer},
  volume = {18},
  pages = {490--502},
  year = {1964},
  doi = {10.1038/bjc.1964.55}
}

@article{norton1988gompertzian,
  author = {Norton, Larry},
  title = {A Gompertzian model of human breast cancer growth},
  journal = {Cancer Research},
  volume = {48},
  pages = {7067--7071},
  year = {1988}
}

@article{simeoni2004predictive,
  author = {Simeoni, Maurizio and Magni, Paolo and Cammia, Cristina and De Nicolao, Giuseppe and Croci, Valter and Pesenti, Enrico and Germani, Massimo and Poggesi, Ilaria and Rocchetti, Marco},
  title = {Predictive pharmacokinetic-pharmacodynamic modeling of tumor growth kinetics in xenograft models},
  journal = {Cancer Research},
  volume = {64},
  number = {3},
  pages = {1094--1101},
  year = {2004},
  doi = {10.1158/0008-5472.CAN-03-2524}
}

@inproceedings{maddox2019simple,
  author = {Maddox, Wesley J. and Izmailov, Pavel and Garipov, Timur and Vetrov, Dmitry P. and Wilson, Andrew Gordon},
  title = {A simple baseline for Bayesian uncertainty in deep learning},
  booktitle = {Advances in Neural Information Processing Systems},
  volume = {32},
  pages = {14958--14969},
  year = {2019}
}

@article{mackay1992practical,
  author = {MacKay, David J. C.},
  title = {A practical Bayesian framework for backpropagation networks},
  journal = {Neural Computation},
  volume = {4},
  number = {3},
  pages = {448--472},
  year = {1992},
  doi = {10.1162/neco.1992.4.3.448}
}

@inproceedings{guo2017calibration,
  author = {Guo, Chuan and Pleiss, Geoff and Sun, Yu and Weinberger, Kilian Q.},
  title = {On calibration of modern neural networks},
  booktitle = {Proceedings of the 34th International Conference on Machine Learning (ICML)},
  series = {Proceedings of Machine Learning Research},
  volume = {70},
  pages = {1321--1330},
  year = {2017}
}

@article{lei2018distribution,
  author = {Lei, Jing and G'Sell, Max and Rinaldo, Alessandro and Tibshirani, Ryan J. and Wasserman, Larry},
  title = {Distribution-free predictive inference for regression},
  journal = {Journal of the American Statistical Association},
  volume = {113},
  number = {523},
  pages = {1094--1111},
  year = {2018},
  doi = {10.1080/01621459.2017.1307116}
}

@inproceedings{romano2019conformalized,
  author = {Romano, Yaniv and Patterson, Evan and Candès, Emmanuel},
  title = {Conformalized quantile regression},
  booktitle = {Advances in Neural Information Processing Systems},
  volume = {32},
  pages = {3543--3553},
  year = {2019}
}

@article{gawlikowski2023survey,
  author = {Gawlikowski, Jakob and Tassi, Cedrique Rovile Njieutcheu and Ali, Mohsin and Lee, Jongseok and Humt, Matthias and Feng, Jianxiang and Kruspe, Anna and Triebel, Rudolph and Jung, Peter and Roscher, Ribana and Shahzad, Muhammad and Yang, Wen and Bamler, Richard and Zhu, Xiao Xiang},
  title = {A survey of uncertainty in deep neural networks},
  journal = {Artificial Intelligence Review},
  volume = {56},
  pages = {1513--1589},
  year = {2023},
  doi = {10.1007/s10462-023-10562-9}
}

@article{begoli2019need,
  author = {Begoli, Edmon and Bhattacharya, Tanmoy and Kusnezov, David},
  title = {The need for uncertainty quantification in machine-assisted medical decision making},
  journal = {Nature Machine Intelligence},
  volume = {1},
  pages = {20--23},
  year = {2019},
  doi = {10.1038/s42256-018-0004-1}
}

@article{collins2015tripod,
  author = {Collins, Gary S. and Reitsma, Johannes B. and Altman, Douglas G. and Moons, Karel G. M.},
  title = {Transparent reporting of a multivariable prediction model for individual prognosis or diagnosis ({TRIPOD}): the {TRIPOD} statement},
  journal = {Annals of Internal Medicine},
  volume = {162},
  number = {1},
  pages = {55--63},
  year = {2015},
  doi = {10.7326/M14-0697}
}

@book{rasmussen2006gaussian,
  author = {Rasmussen, Carl Edward and Williams, Christopher K. I.},
  title = {Gaussian Processes for Machine Learning},
  publisher = {MIT Press},
  address = {Cambridge, MA},
  year = {2006},
  doi = {10.7551/mitpress/3206.001.0001}
}

@article{raissi2019physics,
  author = {Raissi, Maziar and Perdikaris, Paris and Karniadakis, George Em},
  title = {Physics-informed neural networks: a deep learning framework for solving forward and inverse problems involving nonlinear partial differential equations},
  journal = {Journal of Computational Physics},
  volume = {378},
  pages = {686--707},
  year = {2019},
  doi = {10.1016/j.jcp.2018.10.045}
}

@article{karniadakis2021physics,
  author = {Karniadakis, George Em and Kevrekidis, Ioannis G. and Lu, Lu and Perdikaris, Paris and Wang, Sifan and Yang, Liu},
  title = {Physics-informed machine learning},
  journal = {Nature Reviews Physics},
  volume = {3},
  pages = {422--440},
  year = {2021},
  doi = {10.1038/s42254-021-00314-5}
}

@article{che2018recurrent,
  author = {Che, Zhengping and Purushotham, Sanjay and Cho, Kyunghyun and Sontag, David and Liu, Yan},
  title = {Recurrent neural networks for multivariate time series with missing values},
  journal = {Scientific Reports},
  volume = {8},
  number = {1},
  pages = {6085},
  year = {2018},
  doi = {10.1038/s41598-018-24271-9}
}

@inproceedings{rubanova2019latent,
  author = {Rubanova, Yulia and Chen, Ricky T. Q. and Duvenaud, David},
  title = {Latent {ODEs} for irregularly-sampled time series},
  booktitle = {Advances in Neural Information Processing Systems},
  volume = {32},
  year = {2019}
}

@article{revel2004two,
  author = {Revel, Marie-Pierre and Bissery, Anne and Bienvenu, Marc and Aycard, Laurence and Lefort, Catherine and Frija, Guy},
  title = {Are two-dimensional {CT} measurements of small noncalcified pulmonary nodules reliable?},
  journal = {Radiology},
  volume = {231},
  number = {2},
  pages = {453--458},
  year = {2004},
  doi = {10.1148/radiol.2312030167}
}

@inproceedings{kidger2020neural,
  author = {Kidger, Patrick and Morrill, James and Foster, James and Lyons, Terry},
  title = {Neural controlled differential equations for irregular time series},
  booktitle = {Advances in Neural Information Processing Systems},
  volume = {33},
  year = {2020}
}

@inproceedings{debrouwer2019gru,
  author = {De Brouwer, Edward and Simm, Jaak and Arany, Adam and Moreau, Yves},
  title = {{GRU-ODE-Bayes}: continuous modeling of sporadically-observed time series},
  booktitle = {Advances in Neural Information Processing Systems},
  volume = {32},
  year = {2019}
}

@inproceedings{kuleshov2018accurate,
  author = {Kuleshov, Volodymyr and Fenner, Nathan and Ermon, Stefano},
  title = {Accurate uncertainties for deep learning using calibrated regression},
  booktitle = {Proceedings of the 35th International Conference on Machine Learning (ICML)},
  series = {Proceedings of Machine Learning Research},
  volume = {80},
  pages = {2796--2804},
  year = {2018}
}

@article{gneiting2007strictly,
  author = {Gneiting, Tilmann and Raftery, Adrian E.},
  title = {Strictly proper scoring rules, prediction, and estimation},
  journal = {Journal of the American Statistical Association},
  volume = {102},
  number = {477},
  pages = {359--378},
  year = {2007},
  doi = {10.1198/016214506000001437}
}

@inproceedings{ovadia2019trust,
  author = {Ovadia, Yaniv and Fertig, Emily and Ren, Jie and Nado, Zachary and Sculley, D. and Nowozin, Sebastian and Dillon, Joshua V. and Lakshminarayanan, Balaji and Snoek, Jasper},
  title = {Can you trust your model's uncertainty? Evaluating predictive uncertainty under dataset shift},
  booktitle = {Advances in Neural Information Processing Systems},
  volume = {32},
  year = {2019}
}

@inproceedings{oakdenrayner2020hidden,
  author = {Oakden-Rayner, Luke and Dunnmon, Jared and Carneiro, Gustavo and Ré, Christopher},
  title = {Hidden stratification causes clinically meaningful failures in machine learning for medical imaging},
  booktitle = {Proceedings of the ACM Conference on Health, Inference, and Learning (CHIL)},
  pages = {151--159},
  year = {2020},
  doi = {10.1145/3368555.3384468}
}

@article{eisenhauer2009recist,
  author = {Eisenhauer, E. A. and Therasse, P. and Bogaerts, J. and Schwartz, L. H. and Sargent, D. and Ford, R. and Dancey, J. and Arbuck, S. and Gwyther, S. and Mooney, M. and Rubinstein, L. and Shankar, L. and Dodd, L. and Kaplan, R. and Lacombe, D. and Verweij, J.},
  title = {New response evaluation criteria in solid tumours: revised {RECIST} guideline (version 1.1)},
  journal = {European Journal of Cancer},
  volume = {45},
  number = {2},
  pages = {228--247},
  year = {2009},
  doi = {10.1016/j.ejca.2008.10.026}
}

@article{tao2022prediction,
  author = {Tao, Guoqiang and Zhu, Lei and Chen, Qian and Yin, Lihua and Li, Ying and Yang, Jian and Ni, Bingbing and Zhang, Zhen and Koo, Cherie W. and Patil, Prashant D. and Chen, Ying and Yu, Hui and Xu, Ying and Ye, Xin},
  title = {Prediction of future imagery of lung nodule as growth modeling with follow-up computed tomography scans using deep learning: a retrospective cohort study},
  journal = {Translational Lung Cancer Research},
  volume = {11},
  number = {2},
  pages = {250--262},
  year = {2022},
  doi = {10.21037/tlcr-22-59}
}

@article{zhang2020spatiotemporal,
  author = {Zhang, Lei and Lu, Le and Wang, Xiaosong and Zhu, R. M. and Bagheri, M. and Summers, Ronald M. and Yao, Jianhua},
  title = {Spatio-temporal convolutional {LSTMs} for tumor growth prediction by learning {4D} longitudinal patient data},
  journal = {IEEE Transactions on Medical Imaging},
  volume = {39},
  number = {4},
  pages = {1114--1126},
  year = {2020},
  doi = {10.1109/TMI.2019.2943841}
}

\end{document}